\documentclass[sigconf]{acmart}
\AtBeginDocument{%
  \providecommand\BibTeX{{%
    \normalfont B\kern-0.5em{\scshape i\kern-0.25em b}\kern-0.8em\TeX}}}

\usepackage{tcolorbox}
\usepackage{multirow}
\usepackage{multicol}
\usepackage{booktabs}
\usepackage{subfigure}
\usepackage{subcaption}
\usepackage{hhline}
\usepackage{color}
\usepackage{wasysym}
\usepackage{pifont}
\usepackage[ruled, vlined]{algorithm2e}
\usepackage{algpseudocode}
\usepackage{colortbl}
\usepackage{enumitem}
\usepackage{adjustbox}
\usepackage{graphicx}

\newcommand{\best}{\cellcolor[HTML]{DED0B6}}
\newcommand{\second}{\cellcolor[HTML]{FAEED1}}

\def\ModelName{\texttt{GraphWiz}}
\def\DatasetName{{GraphInstruct}}
\newcommand{\vpara}[1]{\vspace{0.05in}\noindent\textbf{#1 }}

\copyrightyear{2024} \acmYear{2024} \setcopyright{acmlicensed}\acmConference[KDD '24]{Proceedings of the 30th ACM SIGKDD Conference on Knowledge Discovery and Data Mining}{August 25--29, 2024}{Barcelona, Spain}\acmBooktitle{Proceedings of the 30th ACM SIGKDD Conference on Knowledge Discovery and Data Mining (KDD '24), August 25--29, 2024, Barcelona, Spain}\acmDOI{10.1145/3637528.3672010}\acmISBN{979-8-4007-0490-1/24/08}

\begin{document}
\title{GraphWiz: An Instruction-Following Language Model for \\ Graph Computational Problems}

\setcopyright{acmcopyright}
\copyrightyear{2024}
\acmYear{2024}

\acmConference[KDD'24]{KDD'24: SIGKDD Conference on Knowledge Discovery and Data Mining}{August 25-29, 2024}{Barcelona, Spain}
\acmBooktitle{KDD'24: SIGKDD Conference on Knowledge Discovery and Data Mining, August 25-29, 2024, Barcelona, Spain}

\makeatletter
\def\authornotetext#1{
	\g@addto@macro\@authornotes{%
	\stepcounter{footnote}\footnotetext{#1}}%
}
\makeatother

\author{Nuo Chen}
\authornote{Authors contributed equally to this research.}
\affiliation{%
\institution{Data Science and Analytics Thrust, The Hong Kong University of Science and Technology (Guangzhou)}
  \country{Guangzhou, China}
  }
\email{chennuo26@gmail.com}

\author{Yuhan Li}
\authornotemark[1]
\affiliation{
\institution{Data Science and Analytics Thrust, The Hong Kong University of Science and Technology (Guangzhou)}
  \country{Guangzhou, China}
  }
\email{yuhanli98@gmail.com}

\author{Jianheng Tang}
\authornotemark[1]
\affiliation{
\institution{The Hong Kong University of Science and Technology\\Data Science and Analytics Thrust, The Hong Kong University of Science and Technology (Guangzhou)}
  \country{Hong Kong SAR, China}
}
\email{jtangbf@connect.ust.hk}

\author{Jia Li}
\authornote{Corresponding author.}

\affiliation{%
\institution{Data Science and Analytics Thrust, The Hong Kong University of Science and Technology (Guangzhou)}
  \country{Guangzhou, China}
}
\email{jialee@ust.hk}

\begin{abstract}

Large language models (LLMs) have achieved impressive success across various domains, but their capability in understanding and resolving complex graph problems is less explored. To bridge this gap, we introduce GraphInstruct, a novel instruction-tuning dataset aimed at enabling language models to tackle a broad spectrum of graph problems through explicit reasoning paths. Utilizing GraphInstruct, we build \ModelName, an open-source language model capable of solving various graph computational problems while generating clear reasoning processes. To further enhance the model's performance and reliability, we integrate the Direct Preference Optimization (DPO) framework within the graph problem-solving context. The improved model, \ModelName-DPO, achieves an average accuracy of 65\% across nine tasks with different complexity levels, surpassing GPT-4 which has an average accuracy of 43.8\%. Our study also investigates the relationship between training data volume and model performance, emphasizing the risk of overfitting as data volume increases. Additionally, we explore the transferability of the proposed model across different tasks and datasets, demonstrating its robust zero-shot generalization capability. \ModelName~offers a new blueprint and valuable insights for developing LLMs specialized in graph reasoning and problem-solving.\footnote{Codes and data are available at \url{https://github.com/nuochenpku/Graph-Reasoning-LLM}}

\end{abstract}

\ccsdesc[500]{Computing methodologies~Artificial intelligence}
\ccsdesc[500]{Mathematics of computing~Graph algorithms}

\keywords{graph algorithms, large language models, instruction tuning}

\maketitle

\section{Introduction}

\begin{figure*}[h]
    \centering
    \includegraphics[width=1\linewidth]{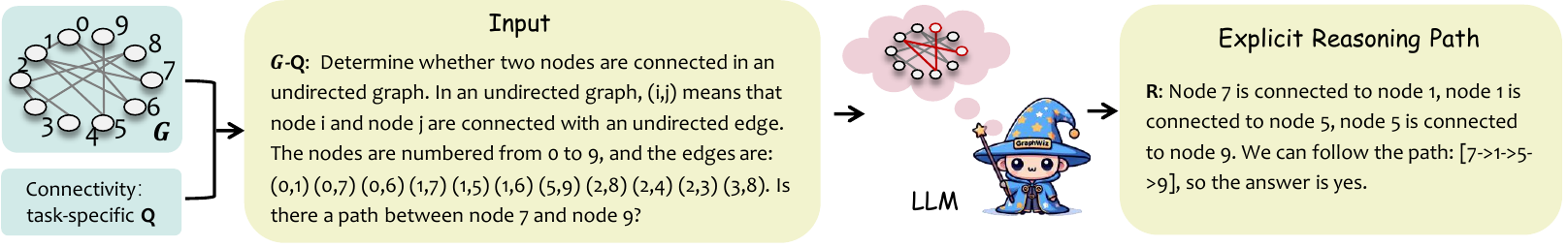}
    \caption{An example of solving the Connectivity task explicitly within natural language via LLMs.}
    \label{fig:question}
\end{figure*}

After witnessing the remarkable capabilities of Large Language Models (LLMs) in text processing, the research community is now exploring their applicability across diverse modalities such as vision, audio, tabular, and spatio-temporal data \cite{kirillov2023segment, scao2022bloom, DBLP:journals/corr/abs-2110-14168, zhou2022least, wei2022chain, selfinstruct}. Graphs, with their non-Euclidean nature and relationship-centric characteristics, present a particularly challenging yet promising frontier for LLM exploration. The synergy between graphs and LLMs has sparked considerable interest due to their potential for bi-directional benefits: integrating graph-based approaches can advance the reasoning abilities of LLMs, enabling them to tackle complex logical tasks such as mathematical problem-solving \cite{chen2023breaking, chen2023good} and commonsense reasoning \cite{yao2022react}. Conversely, LLMs offer powerful capabilities to augment graph analysis, particularly in the realm of semantically-rich, text-attributed networks \cite{fatemi2023talk, wang2023can, tang-etal-2023-fused}.

While the intersection of graphs and LLMs is an emerging field, it remains uncertain \emph{to what extent LLMs can comprehend a graph merely from the context}. Current graph machine learning tasks, such as node classification and link prediction, primarily require LLMs to focus on the semantic aspects of graphs \cite{chen2023exploring,zhao2023graphtext,huang2023can,tang2023graphgpt,chen2023label,ye2023natural,he2023harnessing, wang2024instructgraph, tan2024musegraph}. This often entails a basic understanding of local subgraph structures. Although benchmarks like mathematical problem-solving \cite{trinh2024solving, chen2023breaking, yue2023mammoth, GeoQA} and knowledge-oriented question-answering \cite{ye2023natural,lewisretrieval,chen-etal-2023-orca,you2022end,you2021self} necessitate multi-hop reasoning within a graph, they do not always demand an in-depth understanding of the entire graph's structure. To address this, newer benchmarks like GraphQA \cite{fatemi2023talk, perozzi2024let} and NLGraph \cite{wang2023can} introduce more diverse and complex \textit{graph computational problems} to LLMs, demanding a more profound grasp of graph structures---challenges typically addressed by specific computational algorithms. Figure \ref{fig:question} showcases giving an undirected graph and two nodes in this graph, the LLM needs to answer whether these two nodes are connected through a path. Other similar problems include finding the shortest path and subgraph isomorphism, etc.

Current methodologies for tackling graph problems with language models largely rely on the art of prompt design, seeking to enhance the performance of LLMs through well-crafted prompts \cite{fatemi2023talk, wang2023can}. Notably, this strategy has shown more effectiveness in closed-source LLMs, such as GPT-4 \cite{GPT4OpenAI} and PaLM 2 \cite{anil2023palm}. In contrast, open-source LLMs like LLaMA \cite{touvron2023llama2} demonstrate considerably weaker results using the same methods. Additionally, complex prompts that transform edges into sentences can lengthen the context, restricting model's scalability for larger graphs. Indeed, previous studies \cite{chai2023graphllm, fatemi2023talk, wang2023can, perozzi2024let} have limited graph sizes to very small scales (e.g., fewer than 20 nodes and 100 edges). Existing methods also lack the capability to control or understand the output reasoning process. This raises concerns about whether the model is accurately deducing answers through logical reasoning or simply making correct guesses. \emph{Significantly, there is a noticeable gap in the availability of an open-source language model proficient in explicitly and accurately solving these graph problems.}

In this paper, we aspire to enhance the graph problem-solving abilities of current open-source LLMs and finetune a single model that can solve various types of graph problems and meanwhile output explicit reasoning paths, as the example of connectivity shown in Figure \ref{fig:question}.  However, developing instruction-following models for graph problems is challenging due to the absence of a robust training corpus equipped to aid models in producing explicit reasoning paths.
Such paths are critical for understanding and evaluating the model's reasoning process. To tackle this challenge,  we introduce a novel instruction-tuning dataset named \DatasetName, which is designed to serve as a foundational corpus, enabling language models to not only comprehend graph  problems but also articulate clear and logical reasoning paths. \DatasetName~is constructed with the ``self-augment'' idea: we first sample multiple explicit predictions for each graph problem sample using few-shot Chain-of-Thought (CoT) \cite{wei2022chain} prompting method based on GPT-4, and then finetune a smaller LLM on these predictions. Given a graph problem, the fine-tuned model is used to augment the original corpus by generating diverse reasoning paths. This self-augmentation process is crucial in enhancing the diversity and complexity of the reasoning paths, thereby enriching the dataset and making it more representative of various graph problems. \DatasetName~offers 72,785 training samples across nine graph problem tasks, ranging from linear and polynomial complexity to NP-complete, extending the scope and scale of previous benchmarks.

Upon constructing a corpus specialized for explicit reasoning in graph problems, we fine-tuned current open-source LLMs, including the LLaMA 2 families \cite{touvron2023llama2, touvron2023llama} and Mistral \cite{jiang2023mistral, jiang2024mixtral}. The resulting model, \ModelName, achieves superior performances in solving various graph computational problems. Our training strategy involves mixed-task instruction tuning and directly preference optimization (DPO) alignment. This dual-focused approach ensures the model not only imitates the best examples but also actively avoids common errors. As a result, the most advanced version of \ModelName-DPO achieves an average accuracy of  65\% across all tasks, significantly outperforming GPT-4, which has an average accuracy of 43.8\%. Finally, we delve into the nuanced relationship between training volume, model performance, and overfitting risks. We find that while increasing training data volume generally leads to improved model performance, there is a threshold beyond which the benefits diminish and the risk of overfitting becomes pronounced. Additionally, we explore the potential of transferring the model’s reasoning ability across different graph computational problems.

The contributions of this paper are summarized as follows:

\begin{itemize}[leftmargin=*]
\item We collect the first large-scale instruction-tuning dataset named \DatasetName~specifically designed for training LLMs on a variety of graph computational tasks. This dataset enables the trained models to output explicit reasoning paths and arrive at final answers, significantly enhancing the models' interpretability.
\item We introduce \ModelName, the first open-source LLM specialized for solving graph problems of various types and scales through explicit reasoning. This model markedly outperforms the current best closed-source model, GPT-4, demonstrating superior capability in understanding complex graph structures and addressing related tasks.
\item We analyze potential factors that impact model performance in detail, such as the volume of training data and the sampling strategy for dispreferred samples within the DPO framework. This analysis provides valuable insights into model optimization and performance enhancement across diverse graph tasks.
\end{itemize}

\section{Preliminary and Related Works}


\vpara{Instruction Tuning} Instruction Tuning \cite{ouyang2022training, alpaca, liu2023visual,wei2022finetuned,chen2023breaking} is a process where LLMs are explicitly trained to understand and execute instructions provided in natural language. This approach leverages the models' inherent ability to process and generate language, refining it to follow specified tasks more accurately. Concretely, it involves providing a model with a set of instructions or prompts that define specific tasks it should perform. These instructions are formulated in natural language and are designed to guide the model in understanding and executing various tasks. The process then involves adjusting the model's parameters to optimize its performance on these tasks. 
To date, it's recognized as an essential approach to enhance the LLMs' ability to understand and execute a broad range of tasks based on natural language instructions. In this work:
\begin{tcolorbox}
We aim at leveraging instruction-tuning to build a powerful instruction-following LLM that can understand textural descriptions of graph structures and solve various graph problems explicitly in natural language.
\end{tcolorbox}
In a general scenario, consider an undirected or directed graph $\mathcal{G}=\{\mathcal{V},\mathcal{E}\}$, where $\mathcal{V}$ and $\mathcal{E}$ represent the set of nodes and edges, respectively. Each graph is paired with a task-specific instruction or prompt question $Q$. Let $\mathcal{M}$ denote the current generative LLM, which takes each $\mathcal{G}$-$Q$ pair as inputs and returns the step-by-step textual reasoning path $\mathcal{R}$, as illustrated in Figure \ref{fig:question}. Formally, $\mathcal{R}$ = $\mathcal{M}$($\mathcal{G}$,$Q$), where $\mathcal{R}$ consists of multiple tokens or sentences, commonly referred to as the Chain-of-Thought (CoT) in natural language processing. For certain graph problem tasks, additional information is included in $\mathcal{G}$, such as the edge weight $w$ in identifying shortest paths.



\begin{table*}[]
\caption{Overview of nine tasks in our GraphInstruct benchmark with problem definition, time complexity of representative algorithms, graph type (weighted/directed), node size range, and task difficulty. $|\mathcal V|$ and $|\mathcal E|$ indicate the number of nodes and edges in the graph.}
\centering
\begin{adjustbox}{width=0.98\textwidth}
\begin{tabular}{p{0.12\textwidth}p{0.35\textwidth}ccccc}
\toprule
\textbf{Problem}  &  \textbf{Definition} & \textbf{Time Complexity}    & \textbf{Weighted?}        & \textbf{Directed?}     & \textbf{Node Range}   & \textbf{Difficulty}    \\ \midrule

Cycle Detection &  Detect if a given graph $\mathcal{G}$ contains any cycles. & $O(|\mathcal V| + |\mathcal E|)$  & \ding{55} & \ding{55} & [2, 100]  & Easy \\ \midrule

Connectivity &  Assess if two nodes $u$ and $v$ in a given graph $\mathcal{G}$ are connected via a path. & $O(|\mathcal V| + |\mathcal E|)$  & \ding{55} & \ding{55} & [2, 100] & Easy \\ \midrule

Bipartite Graph Check & Judge if a given graph $\mathcal{G}$ is bipartite. & $O(|\mathcal V| + |\mathcal E|)$  & \ding{55} & \ding{51} & [2, 100]  & Easy \\ \midrule

Topological Sort & Find a topological ordering of vertices in a directed acyclic graph $\mathcal{G}$. & $O(|\mathcal V| + |\mathcal E|)$  & \ding{55} & \ding{51}  & [2, 50] & Easy \\ \midrule

Shortest Path & Compute the shortest path between two specific nodes $u$ and $v$ in a given graph $\mathcal{G}$. & $O(|\mathcal E| + |\mathcal V|\text{log}|\mathcal V|)$  & \ding{51} & \ding{55}  & [2, 100] & Medium \\ \midrule

Maximum Triangle Sum & Find the maximum sum of weights for any connected triplet of vertices in a given graph $\mathcal{G}$. & $O(|\mathcal V|^3)$  & \ding{51} & \ding{55}  & [2, 25] & Medium \\ \midrule

Maximum Flow & Calculate the maximum flow from a source node $s$ to a sink node $t$ in a directed graph $\mathcal{G}$. & $O(|\mathcal V|^2\sqrt{|\mathcal E|})$  & \ding{51} & \ding{51}  & [2, 50] & Medium \\ \midrule

Hamilton Path &  Determine if a given graph $\mathcal{G}$ has a Hamiltonian path that visits each vertex exactly once. & NP-Complete  & \ding{55} & \ding{55}  & [2, 50] & Hard \\ \midrule

Subgraph Matching & Verify if there exists a subgraph in $\mathcal{G}$ that is isomorphic to a given graph $\mathcal{G}'$.  & NP-Complete  & \ding{55} & \ding{51}  & [2, 30] & Hard

 \\ \bottomrule
\end{tabular}
\label{graph_problem}
\end{adjustbox}
\end{table*}

\vpara{Directly Prefered Optimization}
Although current LLMs demonstrate remarkable capabilities across a diverse set of tasks due to extensive training datasets and instruction tuning strategies, they are not immune to issues such as generating misleading information and producing offensive content. Reinforcement Learning from Human Feedback (RLHF) \cite{ouyang2022training, zheng2023secrets, schulman2017proximal}, aims to mitigate these issues by leveraging human feedback to guide the model's learning process. Despite the efficacy of RLHF with PPO \cite{schulman2017proximal} in aligning LLMs with diverse human preferences, this approach necessitates the use of four distinct sub-models, which complicates and increases the cost of training. As an alternative, DPO \cite{rafailov2023direct}  suggests distilling a referential \textit{SFT} policy, $r_{\textnormal{ref}}$, by accentuating preference differences, which is a simple approach for policy optimization. The DPO method optimizes the policy $\pi_0$ using a contrasting pair of outputs $(y_u, y_l)$, as described in the equation below:
\begin{equation}
\begin{aligned}
&\mathcal{L}_{DPO}(\pi_0; r_{\textnormal{ref}}) = -\mathbb{E}_{(x,y_u,y_l)\sim D} \\
&\left[ \log \sigma \left( \beta \log \frac{\pi_0(y_u | x)}{\pi_0(y_l | x)} - \beta \log \frac{r_{\textnormal{ref}}(y_u | x)}{r_{\textnormal{ref}}(y_l | x)} \right) \right],
\end{aligned}
\end{equation}

\noindent where $y_u$ is favored over $y_l$, and $\beta$ is a hyperparameter. \textit{Of note, DPO requires a separate training corpus that belongs to a similar domain with $r_{\textnormal{ref}}$}.



\section{Methodology}

\subsection{GraphInstruct Collection}

\begin{figure}[t]
    \centering
    \includegraphics[width=1\linewidth]{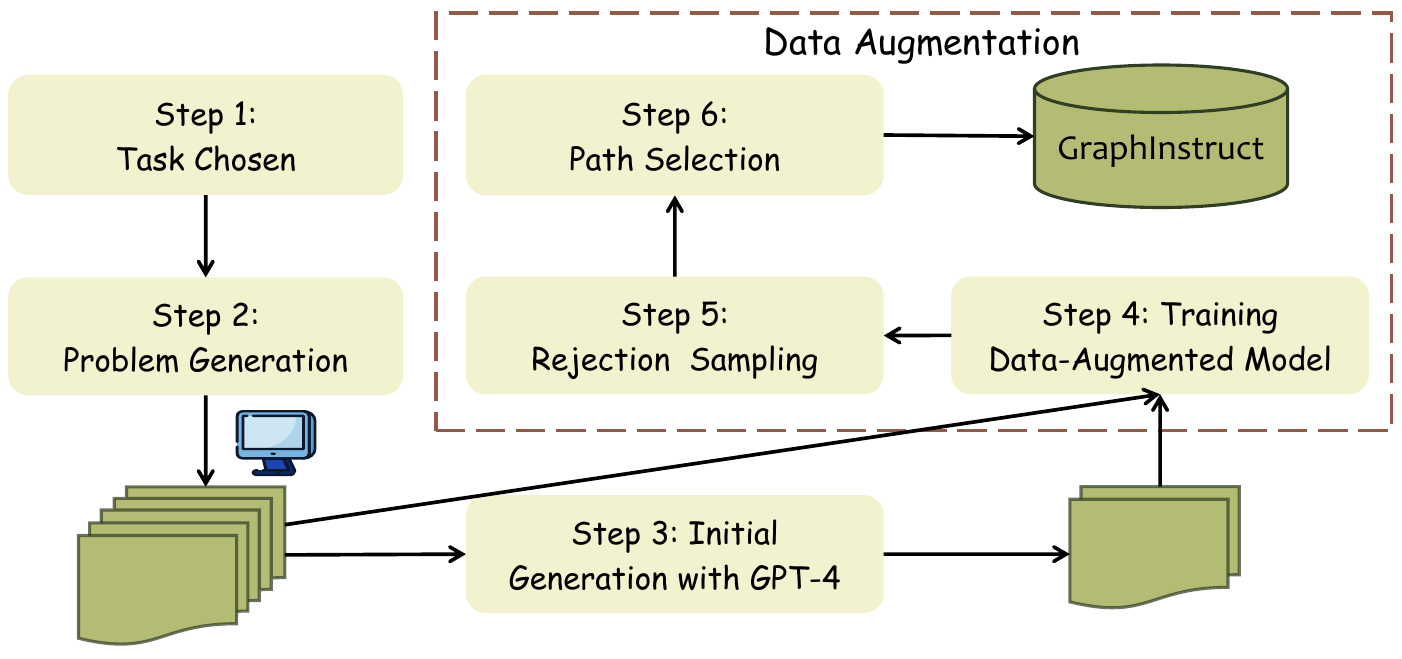}
    \caption{The overview of GraphInstruct collection process.}
    \label{fig:collection}
\end{figure}
\begin{table*}[]
\caption{Statistics of our GraphInstruct corpus, including total samples ($\mathcal{G}$-$Q$), nodes ($\mathcal{V}$) and reasoning paths $\mathcal{R}$. }
\begin{adjustbox}{width=0.9\textwidth}
\small
\begin{tabular}{@{}ll|cccccccccc}
\toprule
&  & \multicolumn{4}{|c|}{\textit{Easy}}                       & \multicolumn{3}{c|}{\textit{Medium}}                   & \multicolumn{2}{c|}{\textit{Hard}}                       &  \\ \cmidrule(l){3-11} 
  &  \multirow{-2}{*}{\textbf{Tasks}} & \multicolumn{1}{|c|}{cycle} & \multicolumn{1}{c|}{connect} & \multicolumn{1}{c|}{bipartite}& \multicolumn{1}{c|}{topology} & \multicolumn{1}{c|}{shortest}   & \multicolumn{1}{c|}{triangle}    & \multicolumn{1}{c|}{flow}   & \multicolumn{1}{c|}{hamilton}   & \multicolumn{1}{c|}{Subgraph} & \multirow{-2}{*}{\textbf{Sum.}}    \\ \midrule

\multirow{3}{*}{\textbf{Train}}&Total $\mathcal{G}$-$Q$ & 3,717 & 2,687 & 2,013 & 902& 1,392&2,756&405&2,097&1,435 &17,158 \\
&Total $\mathcal{V}$ & 84,980 & 79,853 & 58,860 & 10,146& 23,204&14,714&4,333&33,284&7,847 &315,051 \\
&Total $\mathcal{R}$ & 13,122 & 10,001 & 9,324 & 4,481& 5,859&13,483&747&8,454&6,274&72,785 \\
\midrule
\multirow{2}{*}{\textbf{Test}}&Total $\mathcal{G}$-$Q$ & 400 & 400 & 400 & 400& 400&400&400&400&400 &3,600 \\
&Total $\mathcal{V}$ & 19,570 & 19,500 & 19,515 & 9,449& 19,449&4,990&10,024&9,732&6,594 &118,823\\

\bottomrule 

\end{tabular}
\end{adjustbox}
\label{statictics}
\end{table*}

\begin{table}
\caption{Comparison between ours and other typical datasets. }
\label{dataset_compare}
    \centering
    \begin{adjustbox}{width=0.95\columnwidth}
    \begin{tabular}{l|ccccc}
    \toprule
         & Include & Include &  & Node  & Edge  \\
        Datasets & Training Set? & CoTs? & Tasks & Scale & Scale        \\ \midrule
        NLGGraph & No & No & 8 & 9-35 & 10-30 \\ 
        GraphQA & No & No & 12 & 5-20 & 1-100 \\ \midrule
        \textbf{GraphInstruct} & \textbf{Yes} & \textbf{Yes} & \textbf{9} & \textbf{2-100} & \textbf{5-500} \\ \bottomrule
    \end{tabular}
    \end{adjustbox}
    \vspace{-4pt}
\end{table}
In the pursuit of enhancing LLMs' capabilities to tackle graph computational problems with explicit reasoning paths, we develop a new dataset named GraphInstruct. This section outlines the detailed procedures followed to compile and enhance this dataset, including graph task selection, graph problem generation, and subsequent data augmentation. An overview of our dataset collection process is illustrated in Figure \ref{fig:collection}.

\subsubsection{Graph Task Selection.} In this work, we meticulously choose a diverse set of nine graph computational tasks that encompass different levels of computational complexity. We include four linear complexity tasks: \textit{Connectivity}, \textit{Cycle Detection}, \textit{Bipartite Graph Checking}, and \textit{Topological Sort}; three polynomial complexity tasks: \textit{Shortest Path}, \textit{Maximum Triangle Sum}, and \textit{Maximum Flow}; and two NP-Complete tasks: \textit{Hamilton Path} and \textit{Subgraph Matching}. These tasks are defined in Table \ref{graph_problem}, with detailed explanations provided in Appendix \ref{sec:definiton}. These nine graph tasks provide a comprehensive exploration of algorithmic graph theory, allowing us to enhance the theoretical understanding of graph algorithms and address a broad range of practical applications.

\subsubsection{Graph Problem Generation}
To create a diverse and challenging suite of graph problems for model training and testing, we adopt a programming-aid approach inspired by \cite{gao2023pal} that generates random graph problems for each predefined task. Each task is paired with a unique template designed to capture the distinct characteristics of the graphs, such as being directed or undirected and whether the edges contain weights. We employ the Erdős-Rényi (ER) model \cite{erdHos1960evolution} to generate random graphs. Specifically, the ER model accepts two parameters: the number of nodes $n$ and the probability $p$ of an edge existing between any two nodes. For each pair of nodes, the generator randomly decides whether to create an edge between them with probability $p$, resulting in a graph with an average edge density of $p$. We utilize the NetworkX library \cite{hagberg2008exploring} to generate the random graphs and determine the solutions to the graph tasks.

To enhance the quality and diversity of generated problems, we implement specific constraints during the problem generation phase. (1) \textbf{Diverse Distributions}. Each task is crafted to include graph problems with varied distributions. We specify five combinations of node counts and edge densities for each task to introduce varying difficulty levels. The node ranges for each task are detailed in Table \ref{statictics}. It is observed that edge density impacts tasks differently. For example, increasing the edge density in Shortest Path problems generally raises the difficulty by adding more potential routes. Conversely, in Cycle Detection, a higher density may lead to more prevalent cycles, thus simplifying detection. (2) \textbf{Length Constraints}. Considering the token limitations of most current open-source LLMs (e.g., 4096 tokens), we eliminate excessively long graph problems to maintain compatibility, ensuring no graph exceeds 100 nodes. (3) \textbf{Unique Instances}. To build a robust dataset, it is crucial that each problem within the training and testing sets is unique. (4) \textbf{Efficient Graph Description.}  In the textual descriptions of each problem, we only specify the total number of nodes and depict edges using tuples ($u$, $v$), where $u$ and $v$ represent the connected nodes. As illustrated in Figures \ref{fig:question}, \ref{fig:case} and Table \ref{table:Connectivity_prompt}, this graph description language maximize graph sizes within the limits of input length constraints.

We curate an initial set of 27,000 graph problems ($\mathcal{G}$-$Q$) for training, distributed across nine tasks, with 3,000 pairs allocated to each. Furthermore, we prepare 3,600 graph problems ($\mathcal{G}$-$Q$) for testing, also divided among nine tasks, with each task receiving 400 pairs. Notably, we only annotate $\mathcal{R}$ for the training problems.

\subsubsection{Generation of Explicit Reasoning Paths}
\label{path_generation}
The most distinctive aspect of GraphInstruct is that each $\mathcal{G}$-$Q$ pair is coupled with an explicit reasoning path, $\mathcal{R}$. Given the intensive effort required to manually annotate these paths for graph tasks, we utilize GPT-4 to generate preliminary reasoning paths. Each $\mathcal{G}$-$Q$ is fed into GPT-4 with a CoT prompt designed to draw out an explicit reasoning path, concluding with ``The answer is,'' to direct the model towards a definitive solution. To  promote diversity in responses, we employ a sampling strategy with a temperature setting of 0.9, repeating the procedure three times for each problem. We then select only those $\mathcal{G}$-$Q$ pairs that demonstrate accurate reasoning paths---those leading to the correct answer---to form our initial dataset, $D_1$, which includes fewer than 8,000 samples.

\vpara{Data Augmentation with Rejection Sampling.} Observations indicate that GPT-4 struggles with numerous graph tasks, such as yielding fewer than 100 correct samples for the Maximum Flow task in $D_1$. To enrich the dataset with a broader range of reasoning paths, we implement a rejection sampling strategy \cite{chen2023breaking}. Using $D_1$, we develop a specialized data augmentation model, designated as $\mathcal{M}_1$, serving two primary functions: (1)\textbf{ Increasing $\mathcal{G}$-$Q$ quantity}: $\mathcal{M}_1$ is aimed at correctly answering questions that GPT-4 previously mishandled, thus increasing the number of viable $\mathcal{G}$-$Q$ pairs. (2) \textbf{Generating diverse paths}: $\mathcal{M}_1$ is also designed to generate multiple diverse reasoning paths for the same $\mathcal{G}$-$Q$, thereby enhancing the diversity of our dataset. Specifically, Specifically, $\mathcal{M}_1$ processes the initial 27,000 graph problems, with the goal of producing multiple reasoning paths per problem through repeated sampling, aiming to significantly broaden the dataset’s diversity.

\begin{figure*}[t]
    \centering
    \includegraphics[width=1\linewidth]{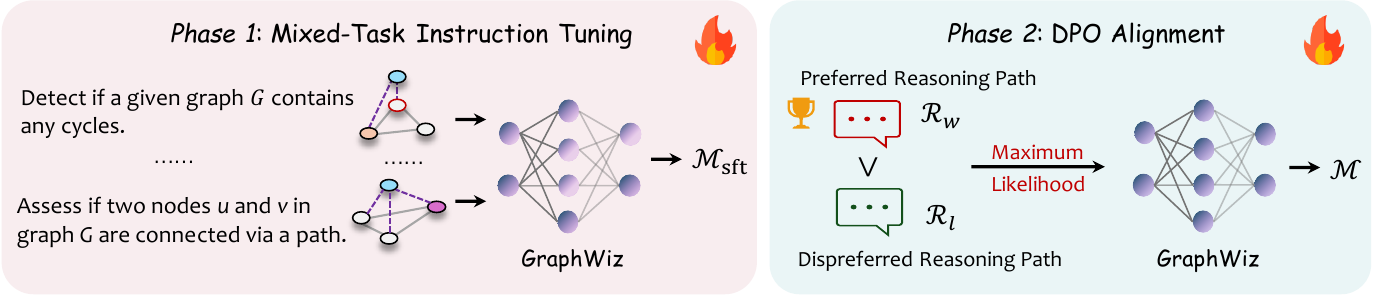}
    \caption{Training pipelines of \ModelName.}
    \label{fig:pipeline}
\end{figure*}

\vpara{Diverse Reasoning Path Selection.} A pivotal aspect of rejection sampling involves selecting correct and diverse reasoning paths to ensure the dataset’s quality and utility. Balancing accuracy with diversity presents significant challenges in this phase. To tackle these, we employ refined strategies divided into string-based and semantic-based approaches for selecting distinct generated reasoning paths:
\begin{itemize}[leftmargin=*,itemsep=0pt,parsep=0pt,topsep=0pt,partopsep=0pt]
    \item \textbf{String-based strategies:} These include using \textit{edit distance}, \textit{TF-IDF} \cite{7754750}, and \textit{Jaccard similarity} to evaluate the uniqueness and relevance of the reasoning paths, ensuring a broad representation of problem-solving techniques.
    \item \textbf{Semantic-based strategies:} We utilize Sentence-BERT \cite{reimers-gurevych-2019-sentence} to encode the reasoning paths and apply \textit{cosine similarity} and \textit{K-means clustering} \cite{kodinariya2013review} to assess their semantic coherence and diversity, encouraging distinct and logically coherent  paths.
\end{itemize}

In our string-based and semantic-based selection strategies, except for K-means clustering, the methodology necessitates a reference or anchor sample to effectively measure the diversity of reasoning paths. This anchor acts as the pivot for assessing the diversity of other paths. We select the \textit{longest correct reasoning path generated for each problem} as the anchor, based on the assumption that it is likely to encompass a comprehensive and detailed reasoning process \cite{jin2024impact}, thus serving as an ideal reference for evaluating the diversity of other paths.

After applying each selection strategy, we compile the final \DatasetName, $D$, by incorporating the reasoning paths that exhibit the greatest diversity compared to the anchor path, as determined by each strategy. In contrast, K-means clustering groups the generated reasoning paths based on their semantic similarity, from which we identify and select a representative path from each cluster, filtering out repetitive reasoning paths and ensuring the left paths remain diverse and distinct. This approach guarantees that \DatasetName~contains not only correct solutions to graph problems but also a broad spectrum of reasoning styles and methodologies. \DatasetName~fosters the development of LLMs capable of generating explicit and interpretable reasoning paths, significantly enhancing their problem-solving proficiency across a diverse array of graph computational problems. Through this endeavor, we aim to bridge the gap between the current capabilities of LLMs and the complex demands of graph problem-solving.

\subsubsection{GraphInstruct Statistics} GraphInstruct encompasses 17,158 graph-question ($\mathcal{G}$-$Q$) pairs, 313,051 nodes ($\mathcal{V}$), and 72,785 reasoning paths ($\mathcal{R}$), with each graph problem potentially linked to multiple paths. Easy tasks are more prevalent, while tasks like Maximum Flow and Topology Sort are less common. This imbalance is primarily due to the difficulty of generating correct reasoning paths, even with advanced models such as GPT-4. Reasoning paths are pivotal in our dataset, as a single $\mathcal{G}$-$Q$ pair may correspond to multiple unique paths, thereby enriching the diversity of GraphInstruct. Additionally, the test set introduces more complex challenges, where an increased number of nodes generally indicates greater difficulty in graph-related problems. This comprehensive dataset forms a robust foundation for training and testing models across a range of graph computational challenges, from basic to advanced levels, as the comparison in Table \ref{dataset_compare}.

\subsection{\ModelName}

Based on the GraphInstruct dataset, we develop \ModelName, which is designed to enhance the capabilities of current LLMs in solving graph computational problems with explicit reasoning paths. The training methodology for \ModelName involves a novel two-phase process. Initially, we employ Mixed-Task Instruction Tuning to refine the model’s ability to interpret and solve a diverse array of graph problems. The subsequent phase, DPO Alignment, further sharpens the model’s reasoning by training it to differentiate between more and less effective problem-solving paths. This approach, especially the application of DPO in the realm of graph problems, marks a significant advancement in teaching LLMs not only to generate explicit answers but also to develop and adhere to a logical reasoning process that mimics expert problem-solving behavior.

\subsubsection{Phase 1: Mixed-Task Instruction Tuning}

The initial phase of our training, called Mixed-Task Instruction Tuning, involves training the model simultaneously on all nine graph tasks from the GraphInstruct dataset. This strategy is designed to provide the model with comprehensive capabilities to solve a diverse array of graph problems by adhering to task-specific instructions. Additionally, it leverages the synergies among different tasks to enhance overall model performance.

During this phase, we employ Supervised Fine-Tuning (SFT) with a conventional language modeling loss. The model is trained to process both the graph $\mathcal{G}$ and its corresponding prompt question $Q$ as inputs and to generate textual reasoning paths $\mathcal{R}$ as outputs. The SFT loss is computed based on the discrepancy between the model’s predicted reasoning paths and the actual reasoning paths in the dataset, formally defined as follows:
\begin{equation}
    \mathcal{L}_{\text{LM}} = -\sum_{i=1}^{N} \sum_{j=1}^{M}\log \text{P}(\mathcal{R}_{i,j} | \mathcal{G}_i, Q_i; \theta)
\end{equation}
where $N$ represents the number of examples, $M$ is the total number of reasoning paths for $\mathcal{G}_i$-$Q_i$, and $\theta$ are the parameters of the model.

This optimization process trains the model to map graph structures and textual prompts to their corresponding reasoning paths, thereby fostering a deep understanding of how to navigate and solve graph problems using explicit reasoning. The model developed in Phase 1 is named $\mathcal{M}(\theta)_{\textnormal{sft}}$.

\subsubsection{Phase 2: DPO Alignment of Reasoning Abilities}

The second phase of training incorporates Direct Preference Optimization (DPO) to enhance the model’s reasoning capabilities. DPO refines the learned SFT model, $\mathcal{M}(\theta)_{\text{\textnormal{sft}}}$, by aligning it more closely with preferred outcomes using a training corpus similar to the SFT data domain. Specifically, DPO employs input pairs labeled ($\mathcal{R}_w, \mathcal{R}_l$), where $\mathcal{R}_w$ and $\mathcal{R}_l$ denote the preferred and less preferred reasoning paths, respectively. In the context of graph problem-solving, the challenge is: \textit{how do we obtain $\mathcal{R}_w$ and $\mathcal{R}_l$ for each $\mathcal{G}$-$Q$ pair?}

To generate these pairs for each $\mathcal{G}$-$Q$, we start by generating a new set of 9,000 graph problems. Using the SFT model $\mathcal{M}(\theta)_{\textnormal{sft}}$, we infer outcomes 20 times for each problem. Consistent with the approach in Section \ref{path_generation}, we select the longest correct path as the preferred reasoning path ($\mathcal{R}_w$). For less preferred paths ($\mathcal{R}_l$), the selection process includes: (1) Using the preferred sample as an anchor, we employ the string-based and semantic-based selection methods (excluding K-means) described in Section \ref{path_generation} to identify the \textit{closest incorrect path} to the preferred one. (2) We then use Majority Voting \cite{wang2022self} to determine the most frequently occurring incorrect path across these strategies, which serves as our dispreferred sample.

This methodology not only ensures the identification of reasoning paths that align with the model’s understanding but also targets hard-negative examples similar to the preferred paths but are fundamentally incorrect. This approach significantly enhances the effectiveness of the DPO stage by rigorously challenging the model to distinguish and preferentially align its reasoning capabilities toward more accurate and logical problem-solving strategies. Finally, we collect 6,166 $\mathcal{G}$-$Q$-$\mathcal{R}_w$-$\mathcal{R}_l$ samples for our DPO training corpus, with detailed statistics available in the Appendix, Table \ref{dpo_statictics}. The formal training objective for DPO is defined as follows:
\begin{equation}
\begin{aligned}
&\mathcal{L}_{DPO}(\mathcal{M}(\theta); \mathcal{M}(\theta)_{\textnormal{sft}}) = -\mathbb{E}_{(x,\mathcal{R}_w,\mathcal{R}_l)\sim D} \\
&\left[  \log \sigma \left( \beta \log \frac{\mathcal{M}(\theta)(\mathcal{R}_w | x)}{\mathcal{M}(\theta)(\mathcal{R}_l | x)} - \beta \log \frac{\mathcal{M}(\theta)_{\textnormal{sft}}(\mathcal{R}_w | x)}{\mathcal{M}(\theta)_{\textnormal{sft}}(\mathcal{R}_l | x)} \right) \right],
\end{aligned}
\end{equation}
where $x$ represents the concatenation of $\mathcal{G}$ and $Q$.

\begin{table*}[]
\caption{Performances of \ModelName~and other baselines on GraphInstruct test set.}
\begin{adjustbox}{width=\textwidth}
\small
\begin{tabular}{@{}llcccccccccc}
\toprule
  & & \multicolumn{4}{|c|}{\textit{Easy}}                       & \multicolumn{3}{c|}{\textit{Medium}}                   & \multicolumn{2}{c|}{\textit{Hard}}                       &  \\ \cmidrule(l){3-11} 
\multirow{-2}{*}{\textbf{Categories}} & \multirow{-2}{*}{\textbf{Algorithms}} & \multicolumn{1}{|c|}{cycle} & \multicolumn{1}{c|}{connect} & \multicolumn{1}{c|}{bipartite}& \multicolumn{1}{c|}{topology} & \multicolumn{1}{c|}{shortest}   & \multicolumn{1}{c|}{triangle}    & \multicolumn{1}{c|}{flow}   & \multicolumn{1}{c|}{hamilton}   & \multicolumn{1}{c|}{subgraph} & \multirow{-2}{*}{\textbf{Average}}    \\ \midrule
 \multirow{3}{*}{Closed-Source}  &        GPT-4 (zero-shot)   & 38.75 &17.00 & 65.25 & 5.00            & 9.25 & 5.75 & 3.25 & 59.25 & 45.50  & 27.67 \\
& GPT-3.5 (2-shot)  &   51.25  &	43.75  &	70.75	& 4.50	& 3.50	 & 17.25  &	8.50  &	54.25 &	43.00  & 32.97      \\ 
& GPT-4 (2-shot)  &   52.50  &	62.75  &	74.25	& 25.25	& 18.25	 & 31.00  &	7.75  &	\second{75.75} &	46.75  & 43.81      \\ \midrule 

 \multirow{3}{*}{Graph Neural Networks}  & 
 GCN&   84.00&  74.00&  82.00&- &5.75&  6.75&   9.25&	-&68.00&-\\
&GIN&	87.50&	73.00&	85.25&-	&7.25&	7.30&	12.00&	 -&66.50&-\\
&GAT&	87.50&	79.25&	85.25&-	&7.25&	7.50&	12.50&	 -&66.25&-\\\midrule 

 \multirow{3}{*}{Mistral-7B}  &  Naive SFT  & 73.75&	83.50 &	78.50&	1.00&	23.00&	47.00&	28.75&	31.75&	41.25&	46.56  \\
 \cmidrule(l){2-12} 

& \textbf{\ModelName} & \second{92.00}&	\best{89.50}&	72.00&	19.00&	\best{31.25}&	38.75&	29.25	&26.50	& \best{85.50}	&53.75 \\

& \textbf{\ModelName-DPO} & 85.50&	79.50&	\second{85.50} &	\best{85.25}&	12.50&	29.00&	35.50&	62.75&	48.50&	58.22   \\ \midrule 

 \multirow{3}{*}{LLaMA 2-7B}  & Naive SFT  &73.75&	83.50 & 41.25 &	4.00 &	9.50& 	30.00 &	16.50 &	69.00 &	75.45 &	44.81  \\ 
 \cmidrule(l){2-12} 
& \textbf{\ModelName} & 91.50& 	87.00 & 	74.00& 	18.00& \second{28.00}& 38.25& 	24.50& 	52.25& 	\second{82.25}& 	55.08   \\ 

& \textbf{\ModelName-DPO} &  89.00 &	82.50&	84.75&	46.75&	24.00 &	\best{52.75}&	\second{43.50}&	\best{81.50}&	77.25&	\best{65.00}   \\ \midrule 


 \multirow{3}{*}{LLaMA 2-13B}  & Naive SFT &  73.75&	83.75&	59.00&	0.50&	11.75&	34.75&	24.25&	59.75&	54.75&	44.69  \\ 
  \cmidrule(l){2-12} 

& \textbf{\ModelName} & \best{94.75}&	87.00&	78.00&	28.00&	27.75&	36.00&	24.50&	59.00&	81.50&	57.39   \\

& \textbf{\ModelName-DPO }& 87.50&	\second{88.50}  &	\best{88.25} &	\second{72.75}&	22.00&	\second{48.75}&	\best{43.75} &	46.50&	77.00&	\second{63.89}  \\ \bottomrule

\label{main_results}
\end{tabular}
\end{adjustbox}
\end{table*}

\begin{table}[]
\caption{\ModelName~performances with increasing ratios of the graph to reasoning path ($\mathcal{G}$:$\mathcal{R}$), which is indicative of the data volume and diversity of reasoning paths available for training. $\mathcal{G}$:$\mathcal{R}$=1:5 means the whole corpus of GraphInstruct.}\label{data_volume}

\begin{tabular}{l|cccc|c}
\toprule
\textbf{Backbone}& {{$\mathcal{G}$:$\mathcal{R}$}} &  \textit{Easy} & \textit{Medium} & \textit{Hard} & \textbf{Average} \\
  
\midrule

 \multirow{5}{*}{{Mistral-7B}} & 1:1 & 61.81 & 32.92 &36.50 & 46.56  \\ 
&1:2 & 68.00 & \second{33.17} & 38.50 & 49.83  \\
& 1:3 & 65.88 & \best{34.25} & 57.50 & 53.47  \\ 
& 1:4 & \second{68.75} & 31.92 & 55.75 & 53.58  \\ 
& 1:5 & 68.13 & 31.50 & 56.00& 53.75 \\
\midrule
 \multirow{5}{*}{{LLaMA 2-7B}} & 1:1 &  58.63 & 28.92 & 38.13 & 44.17  \\ 
 & 1:2 & 68.00 & 26.25 & 61.75& 53.00  \\ 
& 1:3 & \best{69.56} & 31.92 & 60.50 & 54.00  \\ 
& 1:4 & 68.00 & 29.25 & \second{64.00}  & \second{54.36}  \\ 
& 1:5 &67.63 & 30.00 & \best{67.25}  & \best{55.08} \\
\bottomrule
\end{tabular}
\end{table}

\section{Experiments}

In this section, we evaluate the proposed \ModelName~by addressing the following research questions: \textbf{RQ1}: How does \ModelName~perform on these graph tasks, particularly in comparison to the most powerful closed-source model, such as GPT-4? \textbf{RQ2}: What is the impact of training data volume variations on \ModelName’s performance? \textbf{RQ3}: How transferable is \ModelName~across different graph tasks? \textbf{RQ4}: How do changes in the number of nodes in a graph affect \ModelName’s performance, and what is the maximum complexity of graphs it can effectively handle? \textbf{RQ5}: How does hyper-parameter $\beta$ influence model performance.



     

\subsection{Experimental Settings}

Experimentally, we define baselines across two principal settings: (1) \textbf{In-Context Learning}, where the model parameters remain fixed. We utilize state-of-the-art closed-source LLMs as robust baselines, including GPT-3.5 (\texttt{turbo-0613}) and GPT-4 (\texttt{2023-03-14-preview}). Their effectiveness is evaluated using Chain-of-Thought (CoT) prompting, which provides the model with a sequence of examples that sequentially address task solutions. These models are then tasked to replicate this reasoning in their outputs for new problems. We assess their performance under zero-shot and 2-shot scenarios. For zero-shot prompts, the phrase ``Let’s think step by step'' is included to stimulate the generation of autonomous CoTs. (2) \textbf{Training with Naive Labels}: Our focus here is on training smaller-scale LLMs with simplified labels—direct answers like `yes' or `no', without detailed reasoning paths. This approach is termed \textbf{Naive-SFT}. Additionally, we test three Graph Neural Network (GNN) models—Graph Convolutional Network (GCN) \cite{GCN}, Graph Isomorphism Network (GIN) \cite{GIN}, and Graph Attention Network (GAT) \cite{GAT}—on seven of the nine tasks using identical training and test sets as \ModelName. Topology and Hamilton are excluded as they require path-based reasoning, which GNNs are not originally support.

\begin{figure}[t]
    \centering
    \includegraphics[width=1\linewidth]{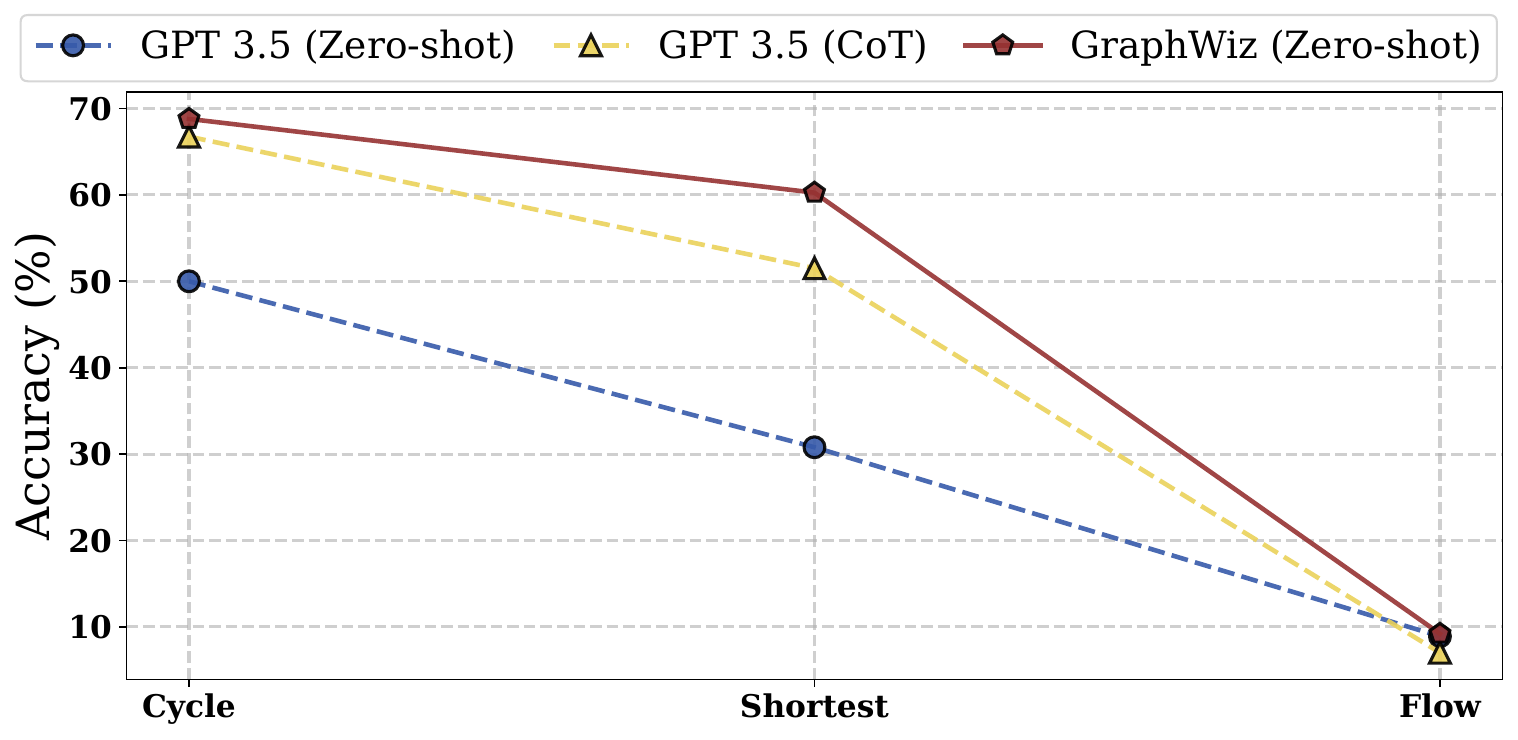}
    \caption{Performance of \ModelName~(Mistral) on NLGraph test set. The results for GPT-3.5 were obtained from its original paper \cite{wang2023can}. In the CoT setting, the cycle task uses 4 exemples, while the shortest and flow tasks use 5 exemples.}
    \label{fig:nlggraph}
    \vspace{-4pt}
\end{figure}

\subsubsection{Implementation} In this work,  we use open-source LLaMA 2-7B/13B and Mistral-7B as backbone models, allowing us to build \ModelName~in multiple scales and architectures. Our codebase is built on DeepSpeed and Huggingface Library. For all models, we set the learning rate, epochs, and max length as 2e-5, 3, and 2048, running on NVIDIA 8*A800 GPUs. The batch size is set to 16 or 32. We use the alpaca-instruction format during training. For the GraphInstruct generation in Section \ref{path_generation}, we use 2-shot CoT prompting for GPT-4. For rejection sampling to augment the data, we train LLaMA 2-13B model for this purpose. Specifically, we sample 30 times with a temperature of 0.9 for each sample, pursuing a high diversity of reasoning paths. For testing, we set temperature as 0 and maximum output tokens as 1024, ensuring the stable and reasonable generation. All GNNs use a hidden dimension size of 16, 2 layers, and 1-dimensional random input features from a Normal distribution.

\subsection{Performance of \ModelName~(RQ1)}
We repeat our evaluation three times and report the average results in Table \ref{main_results} and Figure \ref{fig:nlggraph}. Notably, our models exhibit exceptional results across various backbones, significantly surpassing GPT-4's performance.This consistency is maintained across a range of tasks, from easy to difficult. DPO further enhances the average performance across tasks. However, DPO might adversely affect specific tasks, indicating that while DPO generally improves model reasoning, it may require further tuning to avoid negative impacts on certain problem types. Although Naive SFT shows commendable performance, even slightly better than GPT-4, it is unclear if the outcomes are from random guessing or actual task understanding.

\vpara{Dataset Transfer.} In addition to our constructed test set, we also utilize NLGraph \cite{wang2023can} as part of our testing suite. Given the imbalanced distribution of graph task samples within NLGraph, we select three tasks for evaluation: Cycle Detection, Maximum Flow, and Shortest Path. We directly test our models without any example prompts or further training. Our approach also demonstrates robust cross-dataset transfer capabilities. Figure \ref{fig:nlggraph} illustrates that both \ModelName~maintains a higher level of accuracy compared with GPT-3.5, underscoring the effectiveness of our methodology across datasets with varying distributions. 

 
\vpara{Comparison with GNNs.} GNNs have competitive results on simple graph binary classification tasks like Cycle, Bipartite, and Connectivity. However, as task complexity increases, requiring multi-step numerical reasoning, GNNs' performance markedly declines compared to \ModelName. Beyond the results, we highlight three fundamental differences between GNNs and \ModelName: (1) \textit{Unification}: GNNs require separate training for each task, often necessitating distinct model structures to accommodate different task inputs. In contrast, a single \ModelName~model is capable of handling all tasks effectively. (2) \textit{Generalization}. \ModelName~has the potential to perform zero-shot task transfer: train on some tasks and test on other tasks, a capability absent in GNNs. (3) \textit{Explainability}. Although both models perform similarly on simple tasks such as cycle detection, \ModelName~offers added value by providing step-by-step solutions and identifying specific locations of the cycle, thereby enhancing its explainability. While combining GNNs and LLMs may offer mutual benefits, it also introduces new challenges in model design and joint training. Addressing these challenges in solving graph computational problems remains an area for future work.



\begin{figure}[t]
    \centering
    \includegraphics[width=1\linewidth]{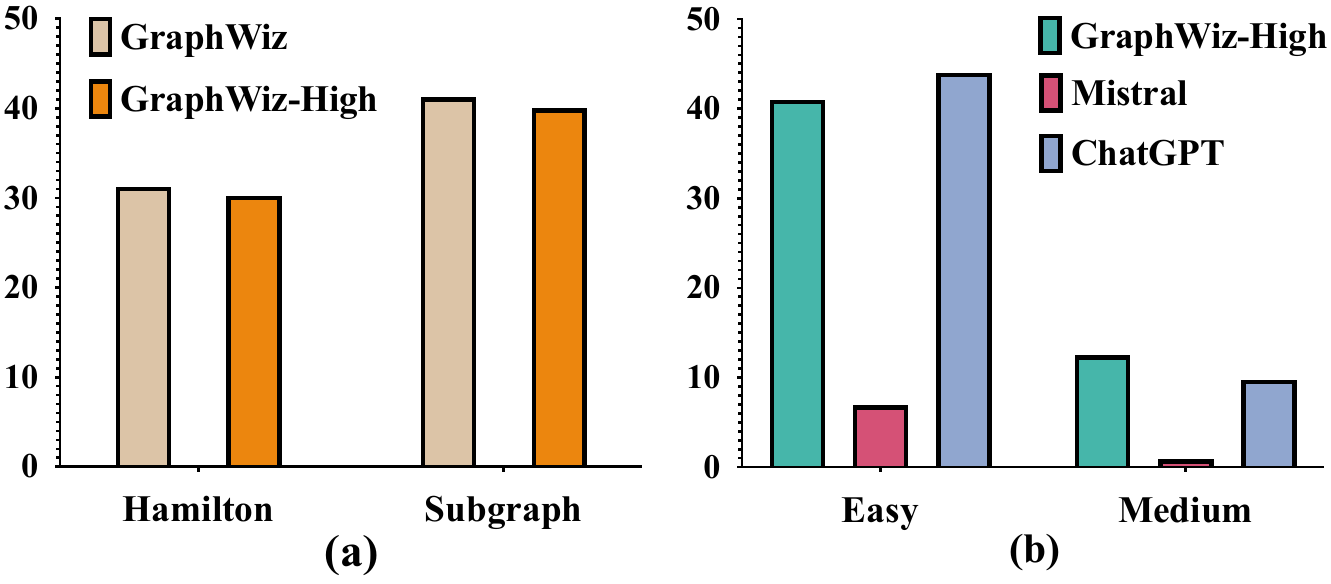}
    \caption{(a) Model performance in two high-complexity tasks; (b) Average model performance of in \textit{easy} and \textit{medium} graph tasks. The backbone of \ModelName~is Mistral.}
    \label{fig:transferability}
\end{figure}



\subsection{Results of \ModelName~with Reasoning Path Increasing (RQ2)}

Table \ref{data_volume} illustrates the average performance of two variants of \ModelName, Mistral 7B and LLaMA 2-7B, across various task categories. Detailed task results are provided in the Appendix, Table \ref{detailed_data_volume}. The performance metrics consider varying ratios of graph to reasoning path ($\mathcal{G}$:$\mathcal{R}$), which represent the data volume and diversity of reasoning paths employed during training. Across all tasks, both models exhibit consistent performance enhancements in correlation with an expanded training corpus. This trend suggests that a greater variety of reasoning paths typically boosts model efficacy across graph-based problems.

Nevertheless, specific tasks such as Triangle Sum and Hamilton Path display negligible or even reduced accuracy improvements with larger data volumes. For instance, the accuracy for \ModelName~(Mistral-7B) in the Triangle Sum task declines from 47.00\% at a 1:1 ratio to 38.75\% at a 1:5 ratio, potentially indicating issues such as overfitting where the model excessively learns specific training data patterns at the expense of generalization to new problems. This observation underscores the importance of careful monitoring and validation during model training to ensure broad applicability and effectiveness.

Furthermore, the $\mathcal{G}$:$\mathcal{R}$ ratios in tasks such as Maximum Flow do not strictly conform to anticipated increments like 1:2 or 1:3. Specifically, in the Maximum Flow task, the GraphInstruct dataset records a $\mathcal{G}$:$\mathcal{R}$ ratio below 1:2 (see Table \ref{statictics}). Despite these deviations, both models can exhibit improved performance in this task, indicating that data volume increases in certain tasks might impact results in others. This suggests the potential for cross-task learning, where models apply reasoning paths learned from one task to enhance performance on another, even without a proportional increase in the data volume for the latter.

\begin{figure}
    \centering
    \includegraphics[width=1.0\linewidth]{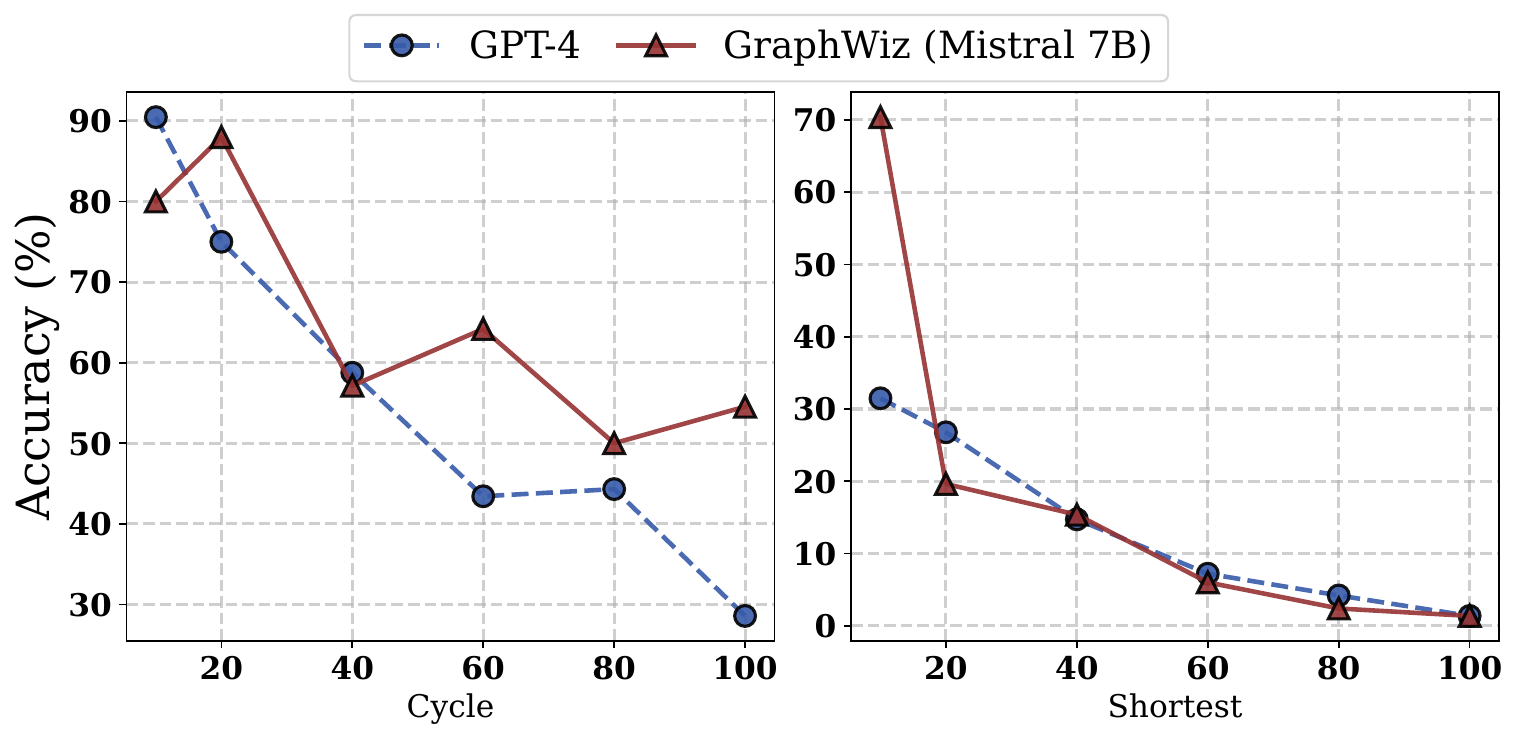}
    \caption{Performance of GPT-4 (2 shot) and \ModelName~based on Mistral 7B with different graph sizes (nodes range) on two tasks. In this work, we use the number of nodes as an indicator of graph complexity. }
    \label{fig:node_range}
    \vspace{-4pt}
\end{figure}

\subsection{Transferability of \ModelName~(RQ3)}
To explore the transferability of \ModelName, we establish an additional model variant to explore this aspect: \ModelName-High. This model is trained exclusively on two high-complexity (NP-Complete) graph tasks: Hamilton Path and Subgraph Matching. To investigate its transferability, we conduct two comparative experiments:

\vpara{Comparison on High-Complexity Tasks.} We first compare \ModelName-High with the regular \ModelName~on high-complexity tasks. Figure \ref{fig:transferability} (a) indicates that \ModelName~performs better, validating the effectiveness of mixed-task training. This outcome also suggests that the model is capable of transferring skills learned from other tasks to specific high-complexity tasks.

\vpara{Zero-Shot Transfer Capability.} We further test the zero-shot transfer capability of \ModelName-High on low and medium-complexity tasks it has never been trained on. As can be seen from Figure \ref{fig:transferability} (b), \ModelName-High shows a significant performance improvement compared to the Mistral-7B backbone. Even when compared to GPT-3.5, our model managed to maintain comparable performance. Considering the vast difference in the number of parameters between GPT-3.5 and \ModelName-High, this indicates that ours possesses a commendable ability to generalize across tasks.

\subsection{Results of \ModelName~with Increasing Problem Complexity (RQ4)}

To address the questions regarding how the model's performance varies with different graph sizes and to figure out the largest graph size the model can effectively solve, we present Figure \ref{fig:node_range} showcasing the performance of the \ModelName~ model on the best-performing task (a) Cycle Detection, and the most challenging tasks (b) Shortest Path. From the figure, we draw the following conclusions: (1) Both \ModelName~and GPT-4 exhibit a decrease in performance as the size of the graph increases. However, our model consistently outperforms GPT-4 across most graph sizes, indicating a more robust understanding and processing of graph structures. (2) We observe that the performance on the latter task, Shortest Path, needs improvement, with a notable decline as the number of nodes increases. This decline can likely be attributed to two main factors: The latter tasks demand high reasoning and memory capabilities as higher time complexity, as well as strong computational skills, which may pose an additional challenge to the models' capacities;  Experimentally, we find that both models primarily rely on enumeration to arrive at solutions. Consequently, as the graph size enlarges, the required enumerative reasoning grows exponentially, leading to a significant drop in accuracy when the number of nodes exceeds 60, after which they exhibit almost no accuracy.

These observations suggest that while \ModelName~shows a clear advantage over GPT-4 in handling graph-related tasks, there is a upper bound of complexity---particularly evident in tasks that require numerical computation---at which the performance of even the most advanced models starts to diminish significantly.

\subsection{Sensitivity Analysis (RQ5)}
\begin{figure}[t]
    \centering
    \includegraphics[width=1\linewidth]{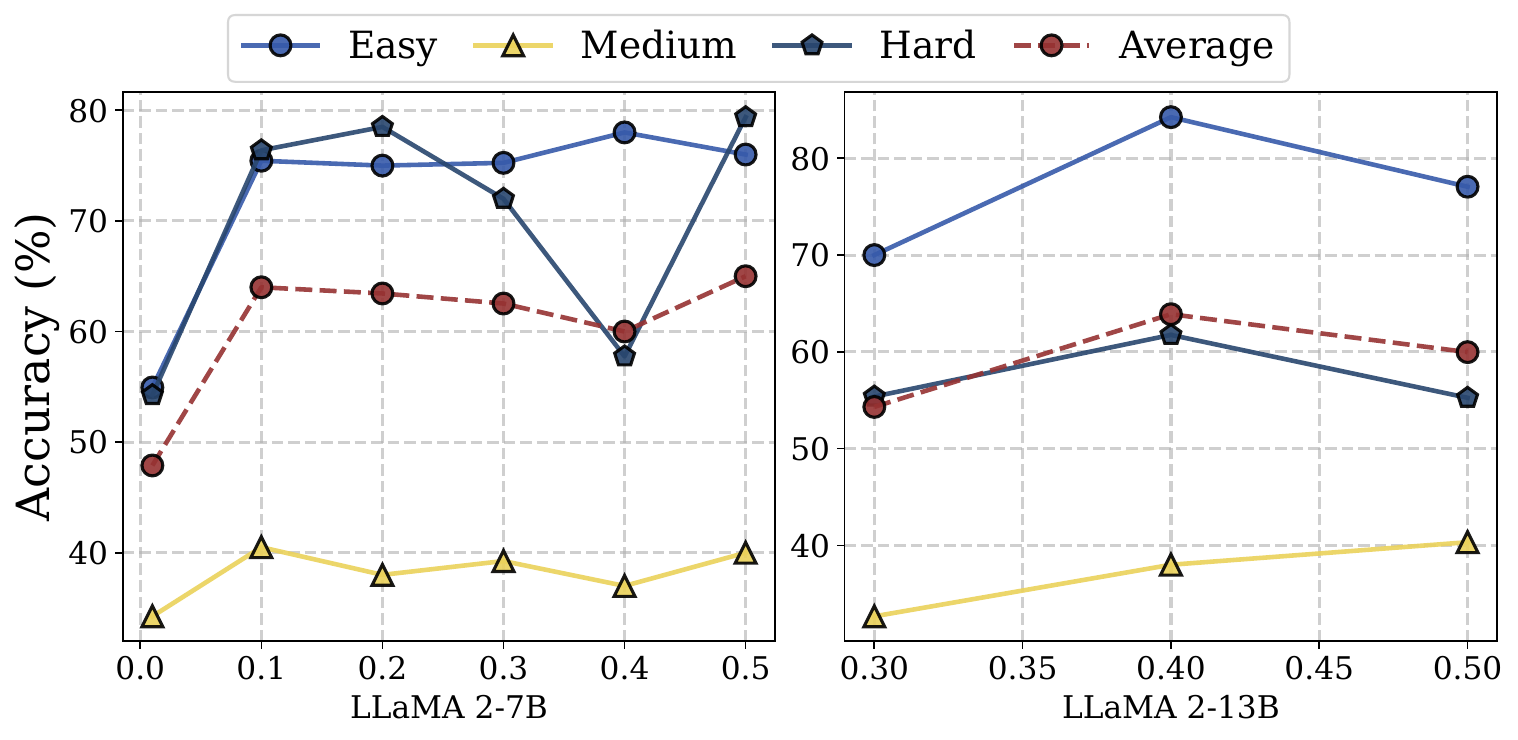}
    \vspace{-10pt}
    \caption{Hyperparameter study of $\beta$.}
    \vspace{-10pt}
    \label{fig:hyper_beta}
\end{figure}

We perform sensitivity analysis to understand the influence of hyperparameter $\beta$ on \ModelName. Figure \ref{fig:hyper_beta} present the average performance of our models based on LLaMA 2-7B and 13B versions in \textit{easy}, \textit{medium} and \textit{hard} tasks. The results indicate that the optimal value of $\beta$ varies depending on the task difficulty and the model size. A higher $\beta$ seems to favor performance on Hard tasks to some extent, but this is not a strictly linear relationship and does not hold consistently across different model sizes. This study highlights the importance of hyperparameter tuning in achieving the best possible model performance across a range of tasks. The analysis suggests that careful tuning of $\beta$ is necessary to achieve the best trade-off performances.

\begin{figure}[t]
    \centering
    \includegraphics[width=1\linewidth]{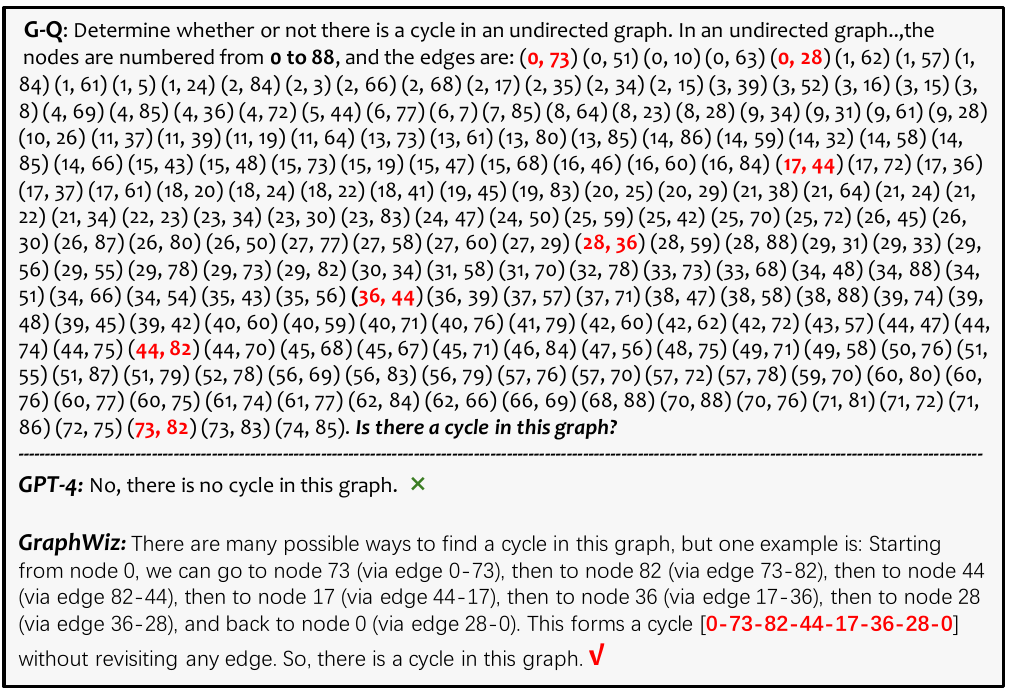}
    \vspace{-10pt}
    \caption{A typical case of \ModelName~and GPT-4. }
    \label{fig:case}
\end{figure}

\subsection{Case Study}
In this subsection, we present a case study of a complex cycle detection problem that involves an undirected graph with \textbf{89 nodes} in Figure \ref{fig:case}, a scale that poses substantial challenges even for human solvers. For a human to detect a cycle in such an extensive graph would require external aids or substantial time, as it is impractical to resolve purely through mental computation. We could observe that GPT-4 outputs a very short incorrect answer, which could arise from the model's constraints in processing long inputs or from an erroneous interpretation of the graph's intricate structure.  This limitation reflects the challenges that conventional LLMs encounter when adapting to graph-theoretic problems. In contrast, \ModelName~correctly detects a cycle, providing a clear and detailed reasoning path. This demonstration of \ModelName's spatial reasoning and memory retention capabilities is significant. It indicates that the model has effectively internalized the principles of graph theory to the extent that it can autonomously navigate through and reason about large-scale and complex graph structures. This case study is a testament to the feasibility of employing \ModelName~ for sophisticated graph problems. 
The model's ability to accurately resolve such complex tasks suggests it has substantial practical applications, offering a robust tool for researchers and practitioners.

\section{Conclusion}
\label{sec:conclusion}

This paper tackles the challenge of equipping LLMs not only to process but to reason explicitly on a variety of graph computational problems. We introduce \DatasetName, a novel instruction-tuning dataset encompassing a broad spectrum of graph problems, designed to enhance the interpretability and reasoning capabilities of LLMs. We subsequently fine-tune open-source LLMs on this dataset, notably from the LLaMA 2 and Mistral series, leading to the creation of \ModelName. This model not only surpasses GPT-4 in performance but also showcases the potential for cross-task transferability of reasoning skills. Our experimental analysis highlights the critical balance needed in training volume to maximize model effectiveness and proposes a promising direction on cross-task transfer learning. Future endeavors will focus on devising more efficient model architectures and training methods to further improve \ModelName’s performance and generalization capabilities.

\begin{acks}
This work was supported by NSFC Grant No. 62206067, HKUST-HKUST(GZ) 20 for 20 Cross-campus Collaborative Research Scheme C019 and Guangzhou-HKUST(GZ) Joint Funding Scheme 2023A03J0673.
\end{acks}

\bibliographystyle{ACM-Reference-Format}
\bibliography{mybib.bib}

\appendix
\section{Task Definition} \label{sec:definiton}

In this section, we provide a detailed definition of each task in our GraphInstruct benchmark. For binary classification tasks, we filter the data to create a balanced set of graphs with both positive and negative labels.
\begin{itemize}[leftmargin=*]
    \item \textbf{Task 1: Cycle Detection.} In an undirected graph $\mathcal{G}=\{\mathcal{V},\mathcal{E}\}$, the task is to detect the existence of a cycle. A cycle can be defined as a sequence of vertices $v_1, v_2, \dots, v_k$ with $k\geq 3$, that forms a closed loop, meaning $v_1=v_k$. Additionally, for all $1\leq i < k$, each vertex $v_i$ must be distinct from the others, and there must be an edge connecting $v_i$ to $v_{i+1}$. 

    \item \textbf{Task 2: Connectivity.} Given an undirected graph $\mathcal{G}=\{\mathcal{V},\mathcal{E}\}$, the task is to assess if two randomly chosen nodes $u$ and $v$ are connected through a sequence of edges.

    \item \textbf{Task 3: Bipartite Graph Check.}  This task is to determine if a directed graph $\mathcal{G}=\{\mathcal{V},\mathcal{E}\}$ is bipartite. A graph is considered bipartite if its nodes can be split into two distinct sets $\mathbf{U}$ and $\mathbf{V}$ such that no two nodes within the same set are adjacent.

    \item \textbf{Task 4: Topological Sort.} The task entails finding a valid topological sort of a directed graph $\mathcal{G}=\{\mathcal{V},\mathcal{E}\}$. In a topological sort, nodes are linearly ordered such that for each directed edge $(u,v)$ from node $u$ to node $v$, the node $u$ is positioned before $v$ in the sequence.

    \item \textbf{Task 5: Shortest Path.} The task requires identifying the shortest path between two nodes in an undirected, weighted graph $\mathcal{G}=\{\mathcal{V}, \mathcal{E}, w\}$, where $w: \mathcal{E} \to \mathbb{R}^+$ assigns a positive weight to each edge. The goal is to find a path connecting the two nodes such that the total sum of the edge weights along this path is minimized.

    \item \textbf{Task 6: Maximum Triangle Sum.} Given an undirected, weighted graph $\mathcal{G} = \{\mathcal{V}, \mathcal{E}, l\} $, where $l: \mathcal{V} \to \mathbb{R}^+$ is a function assigning a positive weight to each node, the task involves finding a triangle, a cycle of three connected vertices $(v_1, v_2, v_3)$, that maximizes the weight sum $l(v_1)+l(v_2)+l(v_3)$.

    \item \textbf{Task 7: Maximum Flow.} Consider a directed, weighted graph $\mathcal{G}=\{\mathcal{V},\mathcal{E}, c\}$, 
    where $c: \mathcal{E} \to \mathbb{R}^+$ is a function assigning a positive capacity to each edge, representing the maximum flow that the edge can support. Given a source node $v_s$ and a sink node $v_t$ in $\mathcal{G}$, the task is to devise a plan to maximize the flow from the source $s$ to the sink $t$. 

    \item \textbf{Task 8: Hamilton Path.} This task is to determine whether there is a Hamilton path in an undirected graph $\mathcal{G}=\{\mathcal{V},\mathcal{E}\}$. A Hamilton path is defined as a path that traverses each node in the graph exactly once.

    \item \textbf{Task 9: Substructure Matching.} Given two graphs, $\mathcal{G}$ and $\mathcal{G}'$, where $\mathcal{G} = \{\mathcal{V}, \mathcal{E}\}$ and $\mathcal{G}' = \{\mathcal{V}', \mathcal{E}'\}$, the task is to determine whether there exists a subgraph within $\mathcal{G}$ that is isomorphic to $\mathcal{G}'$. 

\end{itemize}

\section{Additional Dataset Information}\label{all_prompts}

Table \ref{dpo_statictics} shows the detailed statistics of our DPO training corpus. Table \ref{table:instruction} shows the training and testing instruction of our model. Due to the space limit, we only present the connectivity prompts in Table \ref{table:Connectivity_prompt}. Prompts for other tasks can be found in our GitHub Repository.
\begin{table*}[]
\caption{Statistics of our DPO corpus, including total samples ($\mathcal{G}$-$Q$), nodes ($\mathcal{V}$) and reasoning paths $\mathcal{R}$. }
\begin{adjustbox}{width=0.9\textwidth}
\small
\begin{tabular}{@{}l|cccc|ccc|cc|c}
\toprule
\multirow{2}{*}{\textbf{Tasks}}  & \multicolumn{4}{|c|}{\textit{Easy}}                       & \multicolumn{3}{c|}{\textit{Medium}}                   & \multicolumn{2}{c|}{\textit{Hard}}                       &  \\ \cmidrule(l){2-10} 
     & \multicolumn{1}{|c|}{cycle} & \multicolumn{1}{c|}{connect} & \multicolumn{1}{c|}{bipartite}& \multicolumn{1}{c|}{topology} & \multicolumn{1}{c|}{shortest}   & \multicolumn{1}{c|}{triangle}    & \multicolumn{1}{c|}{flow}   & \multicolumn{1}{c|}{hamilton}   & \multicolumn{1}{c|}{Subgraph} & \multirow{-2}{*}{\textbf{Sum.}}    \\ \midrule

Total $\mathcal{G}$-$Q$-$\mathcal{R}_w$-$\mathcal{R}_l$ & 982 & 972 & 689 & 490& 551&545&405&962&570 &6,166\\
Total $\mathcal{V}$ & 22,132 & 20,705 & 11,546 & 12,802& 21,670&5,842&12,863&24,846
&10,050 &142,456 \\

\bottomrule 

\end{tabular}
\end{adjustbox}
\label{dpo_statictics}
\end{table*}
\begin{table*}[!t]\footnotesize
\centering
\small
\caption{Training and testing prompts of \ModelName~in our experiments.}
\begin{tabular}{l|p{0.8\linewidth}}
\toprule
\multirow{2}{*}{\textbf{ Input Prompts}} &   Below is an instruction that describes a task. \texttt{\string\n}
Write a response that appropriately completes the request. \texttt{\string\n} \texttt{\string\n} \#\#\# Instruction: \texttt{\string\n} \{\textit{Graph-Prompt Question}\}\texttt{\string\n}\texttt{\string\n} \#\#\# Response:   	\\
\bottomrule
\end{tabular}
\label{table:instruction}
\end{table*}

\section{Additional Experiments}
\noindent\textbf{Results on other graph types.} \ModelName~is primarily trained on Erdős-Rényi (ER) graphs due to their representation as general random graphs. To evaluate the generalization ability of \ModelName, we tested it on the Barabási–Albert (BA) model, Path graph, Star graph, and Complete graph. Each graph type included 200 examples for each of the six tasks (Cycle, Connectivity, Bipartite, Hamilton, Shortest, and Flow). The results demonstrate that \ModelName-Mistral-7B achieves an average accuracy of 68.12\% on BA graphs, 54.48\% on Path graphs, 63.17\% on Star graphs, and 83.33\% on Complete graphs, all of which outperform the 47.93\% accuracy on ER graphs. These findings indicate that \ModelName~trained on ER graphs can effectively generalize to other graph types. Notably, \ModelName~achieves 100\% accuracy on the Cycle, Connectivity, and Bipartite tasks for Complete graphs.

\vpara{Effect on the number of reasoning paths.} As shown in Table \ref{data_volume}, Table \ref{detailed_data_volume} provides detailed results of our model’s performance on each task as the ratio of the graph to reasoning paths increases.

\begin{table*}[]
\caption{\ModelName~performances with increasing ratios of the graph to reasoning path ($\mathcal{G}$:$\mathcal{R}$), which is indicative of the data volume and diversity of reasoning paths available for training. $\mathcal{G}$:$\mathcal{R}$=1:5 means the whole corpus of GraphInstruct.}
\begin{adjustbox}{width=0.85\textwidth}
\small
\begin{tabular}{@{}lcccccccccc}
\toprule
  & \multicolumn{4}{|c|}{\textit{Easy}}                       & \multicolumn{3}{c|}{\textit{Medium}}                   & \multicolumn{2}{c|}{\textit{Hard}}                       &  \\ \cmidrule(l){2-10} 
\multirow{-2}{*}{{$\mathcal{G}$:$\mathcal{R}$}} & \multicolumn{1}{|c|}{cycle} & \multicolumn{1}{c|}{connect} & \multicolumn{1}{c|}{bipartite}& \multicolumn{1}{c|}{topology} & \multicolumn{1}{c|}{shortest}   & \multicolumn{1}{c|}{triangle}    & \multicolumn{1}{c|}{flow}   & \multicolumn{1}{c|}{hamilton}   & \multicolumn{1}{c|}{substructure} & \multirow{-2}{*}{\textbf{Avg.}}    \\ 

\midrule

\multicolumn{11}{c}{\ModelName~(Mistral-7B)}  \\ \midrule

 1:1 & 80.25 & 87.50 & \second{78.50} & 1.00 & 23.00 & \best{47.00} & 28.75 & 31.75 & 41.25 & 46.56  \\ 
1:2 & 78.25 & 85.00 & \best{81.75} & \best{27.00} & 28.00 & \best{47.00} & 24.50 & 34.25 & 42.75 & 49.83  \\
1:3 & 90.75 & 88.25 & 62.75 & \second{21.75} & \second{30.75} & 39.75 & \best{32.25} & 30.75 & 84.25 & 53.47  \\ 
1:4 & 90.25 & 87.00 & 78.00 & 19.75 & 26.00 & 43.25 & 26.50 & 26.50 & \second{85.00} & 53.58  \\ 
1:5 & 92.00 & \best{89.50} & 72.00 & 19.00 & \best{31.25} & 38.75 & \second{29.25} & 26.50 & \best{85.50} & 53.75 \\

\midrule

\multicolumn{11}{c}{\ModelName~(LLaMA 2-7B)}  \\ \midrule

 1:1 & 77.75 & 84.00 & 71.75 & 1.00 & 18.50 & \second{46.25} & 22.00 & 31.25 & 45.00 & 44.17  \\ 
1:2 & \best{94.00} & 85.25 & 76.25 & 17.25 & 23.50 & 35.75 & 19.50 & 40.50 & 83.00 & 53.00  \\ 
1:3 & \second{93.00} & \second{89.00} & 72.50 & 18.25 & 24.75 & 40.50 & 22.50 & \second{45.00} & 83.25 & 54.00  \\ 
1:4 & 92.75 & 86.00 & 77.75 & \second{21.75} & 23.25 & 41.50 & 25.25 & 40.25 & 80.75 & \second{54.36}  \\ 

1:5 & 91.50 & 87.00 & 74.00 & 18.00 & 28.00 & 38.25 & 24.50 & \best{52.25} & 82.25 & \best{55.08} \\\bottomrule
\label{detailed_data_volume}

\end{tabular}
\end{adjustbox}
\end{table*}

\begin{table*}[]
\caption{Performance of \ModelName-Mistral-7B on different graph types.}
\label{tab:graph_type}
\begin{tabular}{l|cccccc}
\toprule
  & Cycle & Connectivity & Bipartite & Hamilton & Shortest & Flow \\ \midrule
ER & 92 & 89.5 & 19 & 26.5 & 31.3 & 29.3 \\
BA & 99 & 92 & 68.5 & 51.2 & 59 & 39 \\
Path & 55 & 90 & 34.3 & 77.6 & 37 & 33 \\
Star & 73 & 100 & 21.2 & 44.3 & 73.5 & 67 \\
Complete & 100 & 100 & 100 & 73.9 & 94 & 32.1 \\
\bottomrule
\end{tabular}
\end{table*}

\section{More Cases }
\label{cases}
\begin{figure*}[h]
    \centering
    \includegraphics[width=1\linewidth]{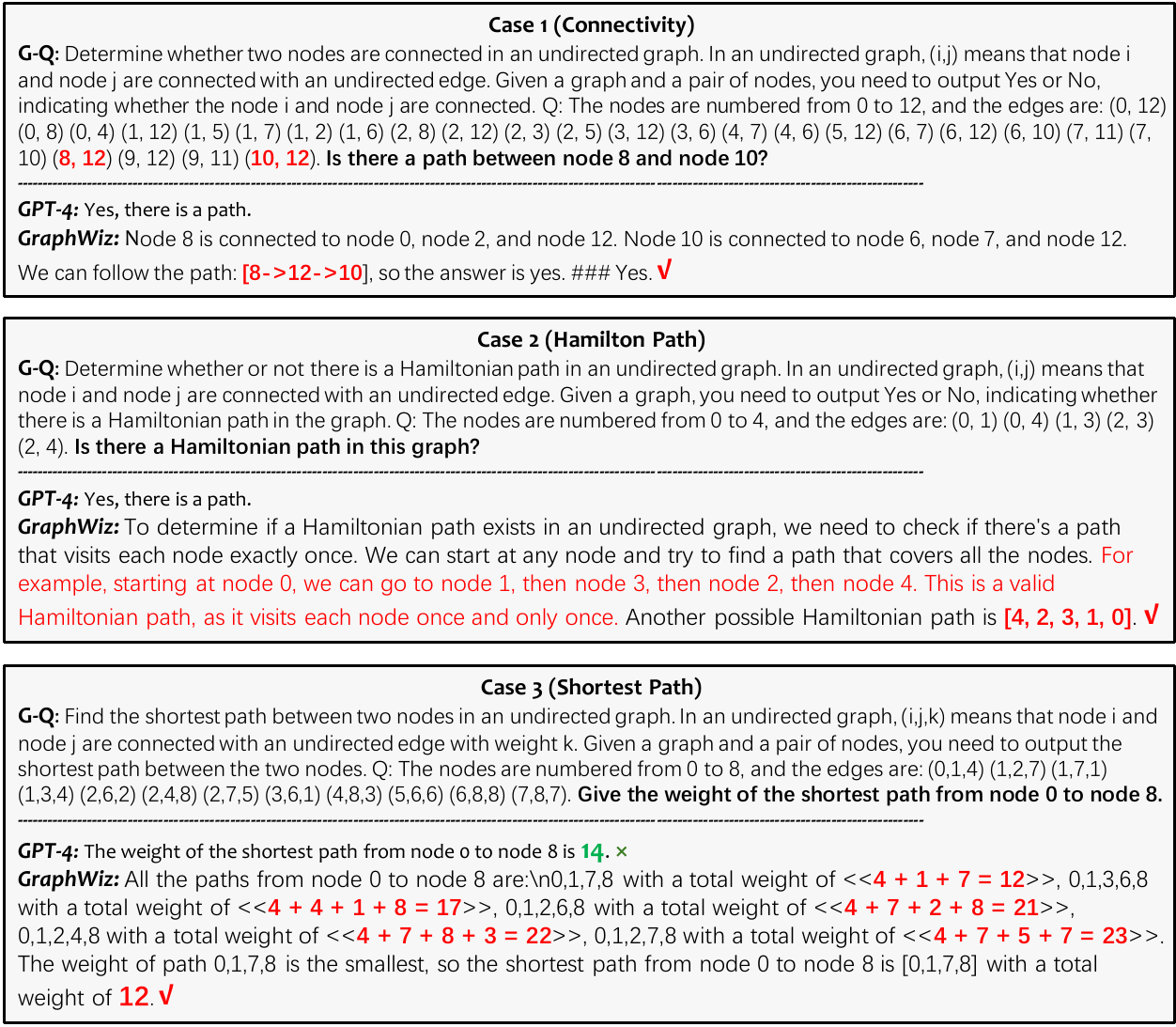}
    \caption{Three cases and responses from GPT-4 and \ModelName.}
    \label{fig:case1}
\end{figure*}

In this section, we present more cases from GPT-4 and our models. In Figure \ref{fig:case1}, we present three cases, encompassing Connectivity, Hamilton Path, and Shortest Path tasks. In case 1 and case 2, we can observe that although GPT-4 outputs the correct answer without any explicit reasoning path while ours not only gives detailed explanations but also drives to the final correct conclusion. Moreover, in case 3, ours even outputs the detailed computation equations for finding the shortest path.

However, we also present a case in the subgraph matching task: Though ours also gives the correct conclusion, there are some errors in the reasoning process, shown in green-colored words. Such phenomenon is called \textbf{Hallucination} \cite{zhang2023siren, ji2023survey} in NLP, which means LLMs sometimes generate texts that contradict facts or depict things that never occurred. We call this case as \textit{false positive response}. This case highlights that hallucination could still be a substantial challenge for current LLMs in graph-problem solving.

Tables \ref{table:Connectivity_prompt},  \ref{table:Cycle_prompt}, \ref{table:Bipartite_prompt}, \ref{table:Topological_prompt}, \ref{table:Shortest_prompt}, \ref{table:Triplet_prompt}, \ref{table:Maximum_prompt}, \ref{table:Hamilton_prompt}, \ref{table:Subgraph_prompt} present the prompts that we used for testing gpt-4 and chatgpt performances in different tasks, which are also employed in Section \ref{path_generation}.

\begin{table*}[]
\centering
\small
\caption{Prompts of Connectivity Task used for GPT-4 and ChatGPT testing, which are also used for initial data generation in Section \ref{path_generation}.} 
\begin{tabular}{p{0.95\linewidth}}
\toprule
\textbf{Prompt of Connectivity Tasks. (2-shot)}
\\
\midrule
Determine if there is a path between two nodes in the graph. \\
Note that (i,j) means that node i and node j are connected with an undirected edge. \\
Given a graph and a pair of nodes, you need to output Yes or No step by step, indicating whether the node i and node j are connected.
\\
Below are several examples:\\
\\

Q: The nodes are numbered from 0 to 5, and the edges are: (0,1) (1,2) (3,4) (4,5). Is there a path between node 1 and node 4?\\
A: Node 1 is in the connected block consisted of node 0, node 1, and node 2.\\
Node 4 is in the connected block consisting of node 3, node 4, and node 5. Node 1 and node 4 are not in the same connected block, so the answer is no. \#\#\# No.\\

Q: The nodes are numbered from 0 to 5, and the edges are: (0,1) (0,2) (1,5) (1,2) (1,3) (2,5). Is there a path between node 2 and node 3?\\
A: Node 2 is connected to node 1, node 1 is connected to node 3. We can follow the path: [2->1->3], so the answer is yes. \#\#\# Yes.

\\
 \texttt{Input:} \\
 \texttt{Q:} \texttt{ <Graph-Prompt Qurestion>}\\
 \texttt{A:}  \\

\bottomrule

\label{table:Connectivity_prompt}
\end{tabular}

\end{table*}

\begin{table*}[!t]
\centering
\small
\caption{ Prompts of Cycle Detection Task used for GPT-4 and ChatGPT testing, which are also used for initial data generation in Section \ref{path_generation}.  } 
\begin{tabular}{p{0.95\linewidth}}
\toprule
\textbf{Prompt of Cycle Detection Task. (2-shot)}
\\
\midrule
Determine whether or not there is a cycle in an undirected graph. Begin with '\#\#\#' to give your final conclusion.\\
In an undirected graph, (i,j) means that node i and node j are connected with an undirected edge.\\
Given a graph, you need to output Yes or No step by step, indicating whether there is a cycle in the graph.\\

Below are examples:
\\
\texttt{two-shot examples}

\\
\\
 \texttt{Input:} \\
 \texttt{Q:} \texttt{ <Graph-Prompt Qurestion>}\\
 \texttt{A:}  \\

\bottomrule

\label{table:Cycle_prompt}
\end{tabular}

\vspace{-5mm}
\end{table*}

\begin{table*}[!t]
\centering
\small
\caption{ Prompts of Bipartite Graph Check Task used for GPT-4 and ChatGPT testing, which are also used for initial data generation in Section \ref{path_generation}.  } 
\begin{tabular}{p{0.95\linewidth}}
\toprule
\textbf{Prompt of Bipartite Graph Check Task. (2-shot)}
\\
\midrule
Determine whether or not a graph is bipartite. \\
In a directed graph, (i->j) means that node i and node j are connected with an directed edge from node i to node j.\\
Given a graph, you need to output 'Yes' or 'No' step by step, indicating whether the graph is bipartite.\\

Blow are examples:\\

\texttt{two-shot examples}

\\
\\
 \texttt{Input:} \\
 \texttt{Q:} \texttt{ <Graph-Prompt Qurestion>}\\
 \texttt{A:}  \\

\bottomrule

\label{table:Bipartite_prompt}
\end{tabular}

\vspace{-5mm}
\end{table*}

\begin{table*}[!t]
\centering
\small
\caption{ Prompts of Topological Sort Task used for GPT-4 and ChatGPT testing, which are also used for initial data generation in Section \ref{path_generation}.  } 
\begin{tabular}{p{0.95\linewidth}}
\toprule
\textbf{Prompt of Topological Sort Task. (2-shot)}
\\
\midrule
Find one of the topology sorting paths of the given graph. \\
In a directed graph, (i->j) means that node i and node j are connected with a directed edge from node i to node j.\\
Given a graph, you need to output one of the topology sorting paths of the graph. \\

Below are several examples:\\

\texttt{two-shot examples}

\\
 \texttt{Input:} \\
 \texttt{Q:} \texttt{ <Graph-Prompt Qurestion>}\\
 \texttt{A:}  \\

\bottomrule

\label{table:Topological_prompt}
\end{tabular}

\vspace{-5mm}
\end{table*}

\begin{table*}[]
\centering
\small
\caption{Prompts of Shortest Path Task used for GPT-4 and ChatGPT testing, which are also used for initial data generation in Section \ref{path_generation}.} 
\begin{tabular}{p{0.95\linewidth}}
\toprule
\textbf{Prompt of Shortest Path Tasks. (2-shot)}
\\
\midrule
Find the shortest path between two nodes in an undirected graph.\\
In an undirected graph, (i,j,k) means that node i and node j are connected with an undirected edge with weight k.\\
Given a graph and a pair of nodes, you need to output the shortest path between the two nodes.\\

Below are several examples:\\
Q: In an undirected graph, the nodes are numbered from 0 to 6, and the edges are: (0,1,1) (1,2,2) (0,2,4) (0,4,2) (2,6,2) (4,6,4) (4,3,5) (6,5,3) (3,5,4). Give the weight of the shortest path from node 0 to node 5.\\
A: All the paths from node 0 to node 5 are:\\
0,2,6,5 with a total weight of <<4 + 2 + 3 = 9>>,\\
0,1,2,6,5 with a total weight of <<1 + 2 + 2 + 3 = 8>>,\\
0,4,6,5 with a total weight of <<2 + 4 + 3 = 9>>,\\
0,4,3,5 with a total weight of <<2 + 5 + 4 = 11>>.\\
The weight of path 0,1,2,6,5 is the smallest, so the shortest path from node 0 to node 5 is [0,1,2,6,5] with a total weight of 8. \#\#\# 8.\\

\\
Q: In an undirected graph, the nodes are numbered from 0 to 4, and the edges are: (0,3,2) (0,4,1) (0,2,1) (4,1,2) (2,1,1) (3,2,4) (2,4,1) (3,4,2). Give the weight of the shortest path from node 3 to node 1.\\
A: All the paths from node 3 to node 1 are:\\
3,2,1 with a total weight of <<4 + 1 = 5>>,\\
3,2,4,1 with a total weight of <<4 + 1 + 2 = 7>>,\\
3,4,1 with a total weight of <<2 + 2 = 4>>,\\
3,4,2,1 with a total weight of <<2 + 1 + 1 = 4>>,\\
3,0,4,1 with a total weight of <<2 + 1 + 2 = 5>>,\\
3,0,2,1 with a total weight of <<2 + 1 + 1 = 4>>,\\
3,4,2,4,1 with a total weight of <<2 + 1 + 1 + 2 = 6>>.\\
The weight of path 3,4,1 is the smallest, so the shortest path from node 3 to node 1 is [3,4,1] with a total weight of 4. \#\#\# 4.\\
\\
 \texttt{Input:} \\
 \texttt{Q:} \texttt{ <Graph-Prompt Qurestion>}\\
 \texttt{A:}  \\

\bottomrule

\label{table:Shortest_prompt}
\end{tabular}

\vspace{-5mm}
\end{table*}

\begin{table*}[!t]
\centering
\small
\caption{ Prompts of Maximum Flow Task used for GPT-4 and ChatGPT testing, which are also used for initial data generation in Section \ref{path_generation}.  } 
\begin{tabular}{p{0.95\linewidth}}
\toprule
\textbf{Prompt of Maximum Flow Task. (2-shot)}
\\
\midrule
Find the maximum flow between two nodes in a directed graph. \\
In a directed graph, (i->j,k) means that node i and node j are connected with an directed edge from node i to node j with weight k. \\
Given a graph and a pair of nodes, you need to output the maximum flow between the two nodes. \\

Below are examples:\\

Q: The nodes are numbered from 0 to 8, and the edges are: (0->2,3) (0->1,9) (0->5,4) (0->3,1) (1->2,7) (1->3,4) (1->5,7) (1->4,5) (2->3,2) (2->5,3) (2->8,2) (2->7,6) (3->5,8) (3->8,4) (3->4,9) (4->7,4) (4->5,6) (4->6,1) (5->6,2) (6->7,6). What is the maximum flow from node 0 to node 2?\\
A: Initially, we can direct a flow of 3 units straight from node 0 to node 2 through the edge (0->2).\\
Further examination reveals that an additional flow can be routed through node 1: the edge (0->1) can carry up to 9 units, and from node 1 to node 2, we can direct 7 units, as limited by the edge (1->2). \\
Summing these flows, we find that a direct flow of 3 units and an indirect flow of 7 units via node 1 give us a total maximum flow of 10 units from node 0 to node 2. \\
This calculation takes into account the various paths and their capacities, ensuring that the flow through any edge does not exceed its capacity. \\
Hence, in this graph, the maximum flow from node 0 to node 2 is 10 units. \#\#\# 10.\\

\\
Q: The nodes are numbered from 0 to 7, and the edges are: (0->3,1) (0->6,5) (0->1,8) (0->5,4) (1->7,1) (1->6,2) (1->2,7) (2->4,5) (2->5,3) (2->3,7) (2->7,4) (3->6,7) (3->5,3) (3->7,7) (4->7,7) (5->7,7) (5->6,1) (6->7,2). What is the maximum flow from node 2 to node 6?\\
A: The graph contains edges like (2->3,7) and (3->6,7), which are crucial for determining the flow. \\
Firstly, there is no direct path from node 2 to node 6, so we explore indirect routes. \\
One such path is through node 3, where node 2 can send a maximum of 7 units to node 3, which in turn can forward up to 7 units to node 6. 
Another route is via node 5; node 2 can send 3 units to node 5, but due to the limited capacity of 1 unit on the edge from node 5 to node 6, only 1 unit can reach node 6 through this path. \\
There's also a path from node 2 to node 7 with a capacity of 4 units, but it doesn't lead to node 6. \\
Thus, by summing the feasible flows, we find that the maximum flow from node 2 to node 6 is 8 units. \#\#\# 8. \\
\\
 \texttt{Input:} \\
 \texttt{Q:} \texttt{ <Graph-Prompt Qurestion>}\\
 \texttt{A:}  \\

\bottomrule

\label{table:Maximum_prompt}
\end{tabular}

\vspace{-5mm}
\end{table*}

\begin{table*}[!t]
\centering
\small
\caption{ Prompts of Maximum
Triangle Sum Task used for GPT-4 and ChatGPT testing, which are also used for initial data generation in Section \ref{path_generation}.  } 
\begin{tabular}{p{0.95\linewidth}}
\toprule
\textbf{Prompt of Maximum
Triplet Sum Task. (2-shot)}
\\
\midrule
Find the maximum sum of the weights of three interconnected nodes. \\
In an undirected graph, [i, k] means that node i has the weight k. (i,j) means that node i and node j are connected with an undirected edge.\\
Given a graph, you need to output the maximum sum of the weights of three interconnected nodes. \\
\\
Below are several examples:
\\

Q: The nodes are numbered from 0 to 4, weights of nodes are: [0, 2] [1, 9] [2, 6] [3, 10] [4, 4], and the edges are: (0, 1) (0, 3) (1, 3) (2, 4) (3, 4). What is the maximum sum of the weights of three interconnected nodes?\\
A: The nodes and their weights are as follows: Node 0 with weight 2, Node 1 with weight 9, Node 2 with weight 6, Node 3 with weight 10, and Node 4 with weight 4. \\
Upon examining the connections between these nodes, it becomes evident that only Nodes 0, 1, and 3 form a fully interconnected set, with each node directly connected to the other two. The sum of their weights is <<2 (Node 0) + 9 (Node 1) + 10 (Node 3) = 21>>.\\
Therefore, the maximum sum of the weights of three interconnected nodes in this graph is 21. \#\#\# 21.\\

\\
Q: The nodes are numbered from 0 to 4, weights of nodes are: [0, 9] [1, 3] [2, 5] [3, 9] [4, 4], and the edges are: (0, 4) (0, 1) (1, 4) (2, 3). What is the maximum sum of the weights of three interconnected nodes?\\
A: The graph comprises nodes 0 to 4, each with respective weights of 9, 3, 5, 9, and 4. \\
Analyzing the graph's edges reveals that Nodes 0, 1, and 4 are the only trio of connected nodes, linked through the edges (0, 4), (0, 1), and (1, 4). \\
By adding their weights:  <<9 (Node 0) + 3 (Node 1) +  4 (Node 4) = 16>>. There are no other groups of three interconnected nodes in this graph. \\
Therefore, the maximum sum of the weights of three connected nodes in this graph is determined to be 16. \#\#\# 16. \\
\\
 \texttt{Input:} \\
 \texttt{Q:} \texttt{ <Graph-Prompt Qurestion>}\\
 \texttt{A:}  \\

\bottomrule

\label{table:Triplet_prompt}
\end{tabular}

\vspace{-5mm}
\end{table*}

\begin{table*}[!t]
\centering
\small
\caption{ Prompts of Subgraph Matching Task  used for GPT-4 and ChatGPT testing, which are also used for initial data generation in Section \ref{path_generation}. } 
\begin{tabular}{p{0.95\linewidth}}
\toprule
\textbf{Prompt of Subgraph Matching Tasks. (2-shot)}
\\
\midrule
Determine if a smaller graph is present as an exact match within a larger graph. \\
In a directed graph, (i->j) means that node i and node j are connected with a directed edge from node i to node j. \\
Given a graph G and a subgraph G', you need to output Yes or No, indicating whether subgraph G' is present within the directed graph G.\\

Below are examples:\\

Q: The nodes of graph G are numbered from 0 to 7, and the edges are: (0->4) (0->5) (0->2) (0->3) (0->1) (0->7) (1->6) (1->5) (1->4) (1->7) (1->3) (2->7) (2->5) (2->6) (2->3) (3->4) (3->6) (3->7) (3->5) (4->7) (4->6) (4->5) (5->6) (5->7) (6->7). The nodes of subgraph G' are numbered from a to e, and the edges are: (a->b) (b->c) (b->e) (b->d) (c->e) (c->d). Is subgraph G' present within graph G as a direct substructure?\\
A: To determine if subgraph G' is present within graph G, let's briefly analyze both graphs:\\
Subgraph G' has the following edges: (a->b), (b->c), (b->e), (b->d), (c->e), (c->d). The key node here is 'b', which has outgoing edges to three different nodes: 'c', 'e', and 'd'. Additionally, 'c' has outgoing edges to both 'e' and 'd'.\\
Now let's find a node in graph G with similar outgoing edges:
Node 0 has outgoing edges to many nodes but is not a match since no single node has outgoing edges to three other nodes that also interconnect as required.\\
Node 1 has outgoing edges to '6', '5', '4', and '7' but none of these nodes have the required interconnections to match 'c', 'e', and 'd'.
Node 2 has outgoing edges to '7', '5', '6', and '3', but again, no suitable interconnections.\\
Node 3 has outgoing edges to '4', '6', '7', and '5'. This resembles 'b' in G', but there must be interconnections between the nodes it points to, matching (c->e), (c->d).\\
Node 4 has outgoing edges to '7', '6', and '5'. If node 4 is 'b', then nodes '7', '6', and '5' could be 'c', 'e', and 'd'. Since '7', '6', and '5' are all interconnected, node 4 and its connected nodes match the structure of G'.\\
Thus, the sequence (4->7), (7->6), (7->5), (6->7), (5->7) in G corresponds to the sequence (b->c), (c->e), (c->d), (e->d), (d->e) in G', which means subgraph G' is present as a direct substructure in graph G. \#\#\# Yes.\\

\\
Q: The nodes of graph G are numbered from 0 to 9, and the edges are: (0->6) (0->2) (1->2) (1->7) (1->3) (3->4) (3->8) (3->9) (4->9). The nodes of subgraph G' are numbered from a to d, and the edges are: (a->d) (a->c) (a->b) (b->d) (b->c) (c->d). Is subgraph G' present within graph G as a direct substructure?\\
A:To find if subgraph G' is present in graph G, we look for a node with out-degree of 3 (like 'a' in G'), and among those outgoing connections, we need two nodes with an out-degree of at least 2 (like 'b' and 'c' in G'), which are also connected to each other and to the third node (like 'd' in G').\\
Examining graph G:\\
Node 0 has out-degree 2, not enough to match 'a'.\\
Node 1 has out-degree 3, so it could be 'a', with nodes 2, 7, and 3 potentially being 'b', 'c', and 'd'.\\
Node 3 has out-degree 3, so it could be 'a', with nodes 4, 8, and 9 potentially being 'b', 'c', and 'd'.\\
Now we must check the connections between the potential 'b', 'c', and 'd' nodes:\\
For node 1 as 'a', nodes 2, 7, and 3 do not have the required mutual connections.\\
For node 3 as 'a', nodes 4, 8, and 9 do not have the required mutual connections either, since there's no edge from 4 to 8 or 9 to 8.\\
None of the nodes satisfy the conditions of subgraph G' fully. \#\#\# No.
\\
\\
 \texttt{Input:} \\
 \texttt{Q:} \texttt{ <Graph-Prompt Qurestion>}\\
 \texttt{A:}  \\

\bottomrule

\label{table:Subgraph_prompt}
\end{tabular}

\vspace{-5mm}
\end{table*}

\begin{table*}[!t]
\centering
\small
\caption{ Prompts of Hamilton Path i used for GPT-4 and ChatGPT testing, which are also used for initial data generation in Section \ref{path_generation}. } 
\begin{tabular}{p{0.95\linewidth}}
\toprule
\textbf{Prompt of Hamilton Path Task. (2-shot)}
\\
\midrule
Determine whether or not there is a Hamiltonian path in an undirected graph.\\
In an undirected graph, (i,j) means that node i and node j are connected with an undirected edge.\\
Given a graph, you need to output 'Yes' or 'No', indicating whether there is a Hamiltonian path in the graph. \\
Below are several examples:\\
Q: The nodes are numbered from 0 to 5, and the edges are: (0, 3) (0, 2) (0, 1) (0, 5) (1, 4) (1, 3) (1, 2) (3, 5) (4, 5). Is there a Hamiltonian path in this graph?\\

A: To determine if a Hamiltonian path exists in an undirected graph, we need to check if there's a path that visits each node exactly once. 
Starting at Node 0, we can go to Node 1 (which connects to Nodes 2, 3, 4).\\
From Node 1, moving to Node 4 seems a strategic choice because Node 4 only connects back to Node 1 and to Node 5. After reaching Node 4, we must go to Node 5.\\
From Node 5, we can go to Node 3, as Node 3 connects to Nodes 0 and 1 (which we've visited) and to Node 5.\\
Finally, from Node 3, we can go to Node 2.\\
So, one possible Hamiltonian path is: [0,1,4,5,3,2].\\
Therefore, there is a Hamiltonian path in this graph. \#\#\# Yes, [0,1,4,5,3,2].\\

\\
Q: The nodes are numbered from 0 to 5, and the edges are: (0,2) (0,1) (4,5) (4,3) (4,2) (5,3) (1,4) (2,5). Is there a Hamiltonian path in this graph?\\
A: To determine if a Hamiltonian path exists in an undirected graph, we need to check if there's a path that visits each node exactly once. 
We can start at node 0. As node 0 is connected with ndoe 2, and node 2 is not visited, we can then visit node 2.\\
As node 2 is connected with ndoe 5, and node 5 is not visited, we can then visit node 5.\\
As node 5 is connected with ndoe 3, and node 3 is not visited, we can then visit node 3.\\
As node 3 is connected with ndoe 4, and node 4 is not visited, we can then visit node 4.\\
As node 4 is connected with ndoe 1, and node 1 is not visited, we can then visit node 1.\\
So, one possible Hamiltonian path is: [0,2,5,3,4,1].\\
Therefore, there is a Hamiltonian path in this graph. \#\#\# Yes, [0,2,5,3,4,1]. \\
\\
 \texttt{Input:} \\
 \texttt{Q:} \texttt{ <Graph-Prompt Qurestion>}\\
 \texttt{A:}  \\

\bottomrule

\label{table:Hamilton_prompt}
\end{tabular}

\vspace{-5mm}
\end{table*}

\begin{figure*}[]
    \centering
    \includegraphics[width=1\linewidth]{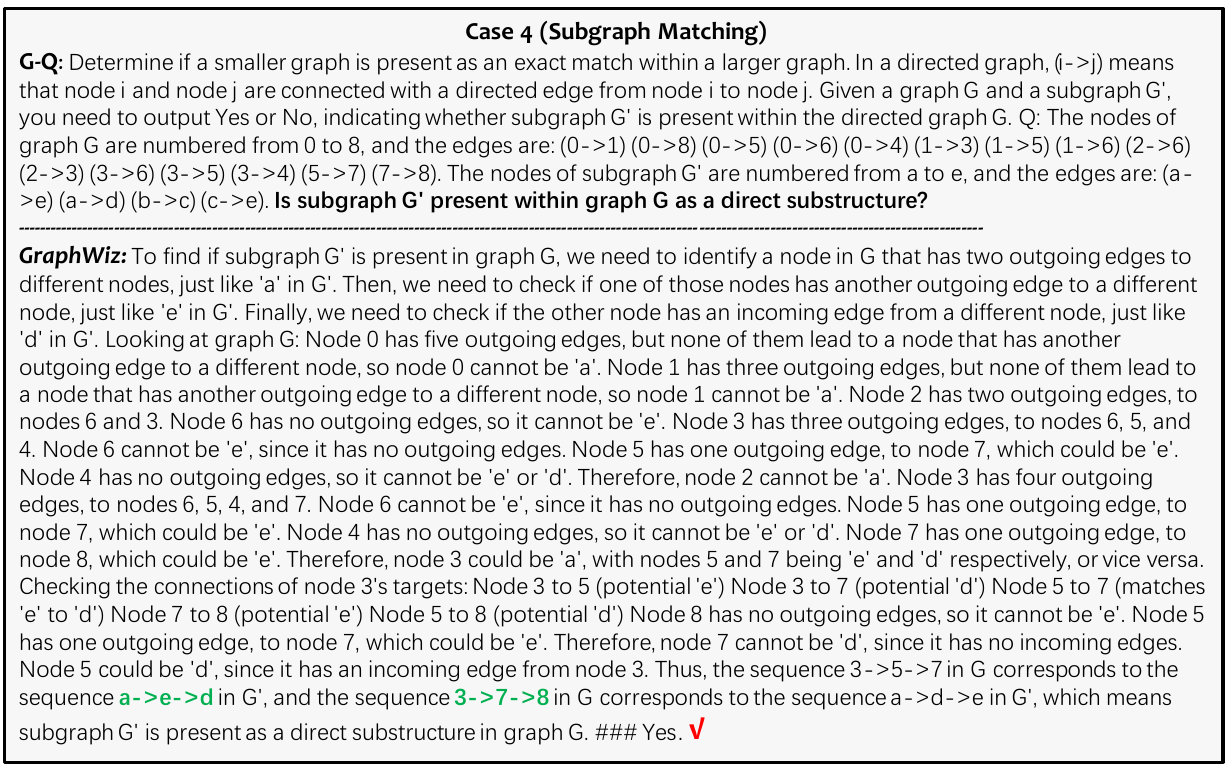}
    \caption{A False positive response from \ModelName.}
    \label{fig:case2}
\end{figure*}
\end{document}